\documentclass[lettersize,journal]{IEEEtran}
\usepackage{amsmath,amsfonts}
\usepackage{algorithmic}
\usepackage{algorithm}
\usepackage{array}
\usepackage[caption=false,font=normalsize,labelfont=sf,textfont=sf]{subfig}
\usepackage{textcomp}
\usepackage{stfloats}
\usepackage{url}
\usepackage{verbatim}
\usepackage{graphicx}
\usepackage{cite}
\usepackage{makecell}

\usepackage{multirow}
\usepackage{amsmath,amssymb,amsfonts}
\usepackage{amsthm}%
\usepackage{mathrsfs}%
\usepackage{xcolor}%
\usepackage{url}
\usepackage{textcomp}
\usepackage{mathrsfs}
\usepackage{booktabs}
\usepackage{color}
\usepackage{bm}
\usepackage{float}
\usepackage{pifont}
\usepackage{enumitem}
\usepackage{booktabs}
\usepackage{arydshln}

\makeatletter
\newcommand*{\new}{\@ifnextchar\bgroup{\new@}{\color{black}}}
\newcommand*{\new@}[1]{{\textcolor{black}{#1}}}
\makeatother

\newcommand{\cmark}{\ding{51}}
\newcommand{\xmark}{\ding{55}}

\begin{document}

\title{ByteNet: Rethinking Multimedia File Fragment Classification through Visual Perspectives}


\author{
  Wenyang Liu, Kejun Wu,~\IEEEmembership{Senior Member,~IEEE}, Tianyi Liu, Yi Wang,~\IEEEmembership{Member,~IEEE}, \\ 
  Kim-Hui Yap,~\IEEEmembership{Senior Member,~IEEE} and Lap-Pui Chau,~\IEEEmembership{Fellow,~IEEE}
  % \thanks{
  % \textcopyright 2020 IEEE.  Personal use of this material is permitted.  Permission from IEEE must be obtained for all other uses, in any current or future media, including reprinting/republishing this material for advertising or promotional purposes, creating new collective works, for resale or redistribution to servers or lists, or reuse of any copyrighted component of this work in other works.}
  % \thanks{
  % This research / project is supported by the National Research Foundation, Singapore, and Cyber Security Agency of Singapore under its National Cybersecurity R\&D Programme (NRF2018NCR-NCR009-0001). 
  \thanks{Wenyang Liu, Tianyi Liu and Kim-Hui Yap are with School of Electrical and Electronics Engineering, Nanyang Technological University, Singapore. (e-mail: wenyang001@e.ntu.edu.sg, liut0038@e.ntu.edu.sg, ekhyap@ntu.edu.sg)}
  \thanks{Kejun Wu is with the School of Electronic Information and Communications, Huazhong University of Science and Technology, Wuhan 430074, China (e-mail: kjwu@hust.edu.cn).}
  \thanks{Lap-Pui Chau and Yi Wang are with the Department of Electrical and Electronic Engineering, The Hong Kong Polytechnic University, Hong Kong. (e-mail: lap-pui.chau@polyu.edu.hk, yi-eie.wang@polyu.edu.hk)}
  \thanks{Corresponding author: Kejun Wu}}

\markboth{Journal of \LaTeX\ Class Files,~Vol.~14, No.~8, August~2021}%
{Shell \MakeLowercase{\textit{et al.}}: A Sample Article Using IEEEtran.cls for IEEE Journals}


\maketitle

\begin{abstract}
Multimedia file fragment classification (MFFC) aims to identify file fragment types, e.g., image/video, audio, and text without system metadata. It is of vital importance in multimedia storage and communication.
Existing MFFC methods typically treat fragments as 1D byte sequences and emphasize the relations between separate bytes (\new{interbytes}) for classification. 
However, the more informative relations inside bytes (\new{intrabytes}) are overlooked and seldom investigated. 
By looking inside bytes, \new{the bit-level details of file fragments can be accessed}, enabling a more accurate classification.
Motivated by this, we first propose \textbf{Byte2Image}, a novel visual representation model that incorporates previously overlooked \new{intrabyte} information into file fragments and reinterprets these fragments as 2D grayscale images. 
This model involves a sliding byte window to reveal the \new{intrabyte} information and a rowwise stacking of \new{intrabyte} n-grams for embedding fragments into a 2D space. Thus, complex \new{interbyte} and \new{intrabyte} correlations can be mined simultaneously using powerful vision networks.
Additionally, we propose an end-to-end dual-branch network \textbf{ByteNet} to enhance robust correlation mining and feature representation. ByteNet makes full use of the raw 1D byte sequence and the converted 2D image through a shallow byte branch feature extraction (BBFE) and a deep image branch feature extraction (IBFE) network. In particular, the BBFE, composed of a single fully-connected layer, adaptively recognizes the co-occurrence \new{of several} some specific bytes within the raw byte sequence, while \new{the} IBFE, built on a vision Transformer, effectively mines the complex \new{interbyte} and \new{intrabyte} correlations from the converted image.
% deep features that existing methods might not consider.
% effective fully connected layer and a sophisticated visual Transformer. 
Experiments on the two representative benchmarks,  including 14 cases, validate that our proposed method outperforms state-of-the-art approaches on different cases by up to 12.2\%.
The code will be released at \url{https://github.com/wenyang001/Byte2Image}.

% exceeds all state-of-the-art comparison methods, achieving an accuracy improvement of 6.5\% in the most challenging scenario. 
%  可以考虑讲讲最高增长多少，或者平均增长多少。你说最有挑战的场景，别人没概念

% especially improving accuracy by 6.5\% on the most complicated scenario.
\end{abstract}


% \begin{highlights}
% \item We are the first ones to consider the bit information within bytes, i.e., \new{intrabytes} information, for file fragment classification.

% \item We propose the Byte2Image technique to treat the byte fragment as a gray 2D image and show the great benefits of doing that.

% \item We get the state-of-the-art results for file fragment classification results.
% \end{highlights}

% % Keywords
% % Each keyword is seperated by \sep
% \begin{keywords}
% File fragment classification \sep 
% Byte2image \sep 
% Byteformer \sep
% Memory forensics \sep
% Vision Transformer \sep
% Data augmentation
% \end{keywords}
% \maketitle

% Main text
\section{Introduction}
\IEEEPARstart{M}{ultimedia} delivers information through diverse forms of media, including but not limited to text, audio, image, and video~\cite{sundararaj2021opposition, cai2024towards, cai2024top, deldjoo2020recommender, wu2024multifocal}, which are combined into a single interactive presentation in a multimedia way; thus it features plentiful interaction among users and devices.
\begin{figure}[htbp]
\centering
\includegraphics[width=3.5in]{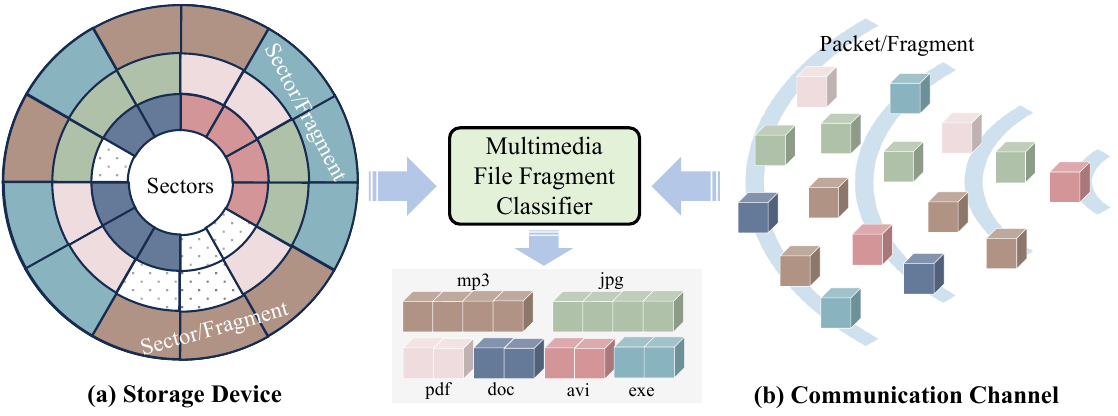} 
\caption{Illustration of storage in sectors and transmission in packets. Multimedia contents saved in a network packet or storage sector can be seen as a fragment.} 
\label{TransmissionStorage}
\vspace{-0.15in}
\end{figure}
The interaction and presentation rely on the transmission and storage of multimedia data. Before multimedia contents are available for users for interaction and presentation, their data are generally delivered remotely from transmission networks or locally from storage media~\cite{ohm2015transmission, liu2024bitstream}. During these processes, data \new{reside} in a network packet or a storage sector \new{as the} basic unit \new{for handling} multimedia file fragments.

\new{An} example of multimedia storage is shown in Fig.~\ref{TransmissionStorage} (a). The space of storage medium is divided into regular sectors, where data need to be broken up into a number of sectors to fit in the available storage space regardless of whether they are in continuous or discontinuous \new{way}~\cite{uzun2015carving}.
%The space of storage mediums is divided into regular sectors, where data is saved in different sectors regardless of whether they are in continuous or discontinuous ways~\cite{uzun2015carving}. 
\new{An} example of multimedia communication is also shown in Fig.~\ref{TransmissionStorage} (b). \new{A bitstream} of data is transmitted through networks by \new{packets}, which \new{are} seen as a single entity for data transmission. Due to the nature of multimedia communication and storage, data \new{are} segmented into fixed-sized units for efficient transmission and reliable storage. Multimedia content carried in a network packet or storage sector can be seen as a ``fragment". 
Multimedia systems generally deliver/store a collection of multiple media sources, e.g., text, audio, image, and video in a fragment-by-fragment manner. 
Therefore, the variety of media types requires multimedia file fragment classification (MFFC), especially in unreliable communication channels and corrupted storage devices for cyber security and digital forensics. An accurate MFFC enables a more resilient multimedia system against channel errors and storage corruption.

Most of the earlier works rely on magic bytes~\cite{pal2009evolution} (e.g., from file headers or footers) or \new{handcrafted} features~\cite{beebe2013sceadan} to identify file types. However, these methods often suffer from limitations such as suboptimal performance and limited flexibility. Recently, with the advances in machine learning, data-driven techniques have shown great promise in MFFC. Compared to the earlier works, they enable implicit feature modeling automatically learned by the neural network, greatly improving the accuracy of file type identification. For example, Haque~\textit{et al.}~\cite{haque2022byte} proposed a Byte2Vec model that used the Skip-Gram model to learn the dense vector representation for bytes in file fragments and then adopted the k-Nearest Neighbor (kNN) algorithm for the file fragment classification. However, since the feature extraction and classification are separated, the overall performance is limited by the individual optimization of each module. To leverage the potential benefits between feature extraction and classification, Mittal~\textit{et al.} proposed an end-to-end method called FiFTy~\cite{mittal2020fifty} that can jointly optimize feature extraction and classification by using a 1D CNN, resulting in superior performance.
These methods typically treat fragments as 1D byte sequences, restricting their ability to explore relations solely between separate bytes (\new{interbyte}). \new{To date}, little attention has been given to the bit information within bytes of fragments, and few studies have emphasized the importance of relations inside a byte (\new{intrabyte}).

\begin{figure}[t]
\centering
\includegraphics[width=3.5in]{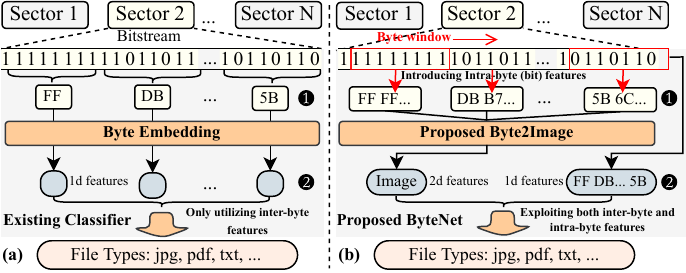}  
\caption{Comparison between existing MFFC methods and our proposed method. (a) Existing methods generally only consider the interbyte features~\cite{mittal2020fifty,haque2022byte}. (b) Our proposed method exploits both interbyte and intrabyte features by two innovations: 1) the intrabyte features are introduced by using a sliding byte window (stride=1 bit) in the proposed Byte2Image, and 2) the raw 1D byte sequence and the converted 2D image are fully leveraged in the proposed ByteNet.}
\label{f:insights}
\vspace{-0.15in}
\end{figure}

\begin{figure}[!t]
\centering
\includegraphics[width=3.5in]{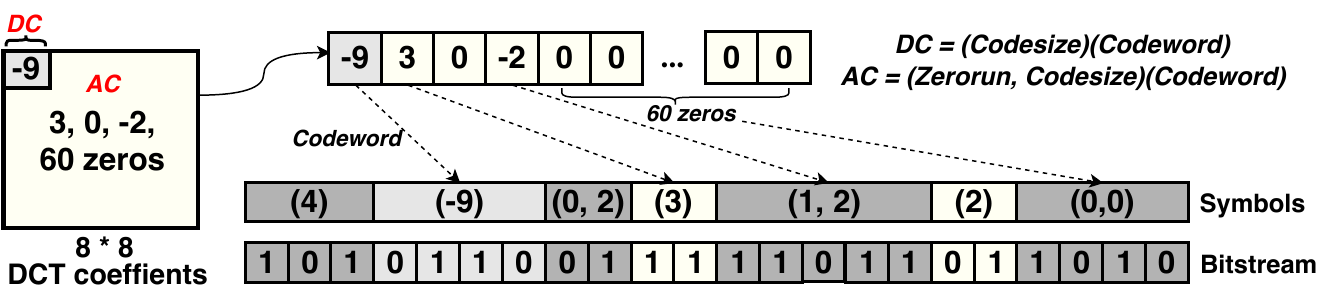}  
\caption{Illustration of Huffman coding used in JPEG files, where intermediate symbols of DCT coefficients are represented as variable-length bits.}
\label{f:jpeg}
\vspace{-0.25in}
\end{figure}

Taking existing works as an example for illustration, as depicted in Fig.~\ref{f:insights} (a), these works treat \new{the} memory sectors of file fragments as byte sequences, where every fixed 8 bits constitute individual bytes (e.g., FF and DB shown in \ding{182}). Their primary focus is on exploring \new{interbyte} relations by learning suitable byte embeddings using data-driven techniques (shown in \ding{183}). 
However, these approaches are inappropriate for classifying files that employ variable-length codes (VLCs) to compress data, such as Huffman codes in JPEG files. 
As Fig.~\ref{f:jpeg} shows, 8*8 DCT coefficients originating from pixel blocks through the Discrete Cosine Transform (DCT) are \new{first} encoded into intermediate symbols such as (4) \new{and} (-9). These symbols are then represented as variable-length bits that may cross byte boundaries. 
In this case, existing methods that rely solely on \new{interbyte} relationships suffer inherent limitations~\cite{horton2023bytes} in exploiting the valuable information inside a byte. 
Thus, it remains challenging to mine the valuable \new{intrabyte} information and consider correlations between \new{interbyte} and \new{intrabyte} relationships, beyond the inherent limitations of 1D byte sequence modeling approaches.

To solve this challenge, we propose a novel visual representation model, \textbf{Byte2Image}, and a specialized classification method, \textbf{ByteNet}. 
The Byte2Image models 1D byte sequences of file fragments \new{via the} 2D representation shown in Fig.~\ref{f:insights} (b).
Specifically, we first employ a byte sliding window (stride = 1 bit) shown in \ding{182} of Fig.~\ref{f:insights} (b) to expose \new{the intrabyte} information as additional bytes. 
However, the length of the augmented byte sequence is dramatically increased, which may potentially introduce lots of extra redundant information. 
To effectively consider both \new{intrabyte} and \new{interbyte} information, we draw inspiration from the effectiveness of CNNs/Transformers~\cite{he2016deep, dosovitskiy2020image} in extracting essential global features while disregarding redundant details for classification. We represent these byte sequences as 2D grayscale images and aim to leverage these networks for classification. Therefore, we represent these byte sequences as 2D grayscale images by stacking these small byte sequence parts row by row and treating each byte as a pixel.
Note that the 1D feature acquisition is consistent with the Fig.~\ref{f:insights} (a).
To ease the performance drop in normal CNNs resulting from the low aspect ratio (width to height) of the constructed image~\cite{ghosh2019reshaping}, we introduce \new{intrabyte} n-grams to make the converted image more square. 
% Finally, inspired from~\cite{cheng2016wide, zheng2017wide}, 
Furthermore, we \new{proposed} a dual-branch network, \new{ByteNet}, in \ding{183} of Fig.~\ref{f:insights} (b) for fragment classification. ByteNet consists of a shallow byte branch feature extraction (BBFE) network and a deep image branch feature extraction (IBFE) network. In particular, the BBFE, composed of a single fully-connected layer, adaptively recognizes the co-occurrence \new{of} some specific bytes within the raw byte sequence, while \new{the} IBFE, built on a vision Transformer, effectively mines the complex \new{interbyte} and \new{intrabyte} correlations from the converted image. 
Thus, the proposed ByteNet can significantly enhance correlation mining and feature representation. 
Our preliminary version of this work~\cite{liu2023byte} is built on the idea of exploiting the valuable \new{intrabyte} information neglected by existing approaches [7, 8] for classification. \new{However,} it only proposes the Byte2Image model to introduce \new{intrabyte} information and view the 1D byte sequence as 2D gray images, while the classification model simply uses the commonly used CNN architectures, e.g., ResNet-18. In this paper, we improve \new{upon} our preliminary work by introducing more image augmentation techniques, developing a novel Transformed-based ByteNet design, and providing an in-depth exploration of the Byte2Image technique along with a comprehensive discussion of the ByteNet design.
The contributions of this work are summarized as follows. 
 \new{intrabytes} correlations by powerful CNNs.

\begin{itemize}
  \item We propose Byte2Image, a novel visual representation model to look inside the bytes of multimedia file fragments for representing fragments in \new{an} \new{interbyte} and \new{intrabyte} manner. It incorporates previously overlooked \new{intrabyte} information into file fragments and reinterprets these fragments as 2D grayscale images. 
  A variety of image augmentation techniques \new{have also been} employed to enhance \new{the} generalization ability and avoid the risk of overfitting.
  \item We propose ByteNet, an end-to-end dual-branch network, to enhance robust correlation mining and feature representation. It fully leverages the raw 1D byte sequence and the converted 2D image through a shallow byte branch feature extraction (BBFE) and a deep image branch feature extraction (IBFE) network, \new{allowing correlations to be mined in both interbyte and intrabyte ways.}
  \item Extensive experiments on two benchmarks validate that our proposed method significantly outperforms state-of-the-art approaches. Ablation studies and visualization demonstrate the effectiveness of our proposed modules.
\end{itemize}

\section{Related Work}

\subsection{Multimedia File Fragment Classification}
Multimedia file fragment classification has undergone extensive study and development over the years. Many earlier works simply rely on magic bytes~\cite{pal2009evolution} (e.g., from file headers or footers) to identify the file type of file fragments. 
Magic bytes are a specific sequence of bytes placed at the beginning or end of a file that allows systems to ensure the \new{integrity of the file}. For instance, in JPEG files, starting file fragments must contain magic byte FFD8. However, these methods usually suffer from significant accuracy drops on seriously fragmented files that may not contain magic bytes. \new{Handcrafted} features, as highlighted in works~\cite{mcdaniel2003content, beebe2013sceadan, li2005fileprints}, offer more reliable performance by considering inherent statistics found in various file types. For example, 
Li~\textit{et al.}~\cite{li2005fileprints} combined the idea of k-means clustering to compute a set of centroid models and used it to find possible file types with improved performances. 
Karresand~\textit{et al.}~\cite{karresand2006oscar} proposed the OSCAR, which also \new{models} the clustering centroids by introducing a new weighted quadratic distance metric between the centroids and the test file fragment sample.

Recently, with the advances in machine learning, learning-based approaches have shown great promise in MFFC. Unlike the laborious task of manually crafting features, learning-based approaches~\cite{wang2018sparse, haque2022byte, mittal2020fifty} leverage neural networks (NNs) to implicitly model features automatically, which has shown significant potential. Wan~\textit{et al.}~\cite{wang2018sparse} used sparse coding to enable automated feature extraction and use a support vector machine (SVM) for classification. Inspired by word embeddings in natural language processing (NLP), Byte2Vec~\cite{haque2022byte} was proposed. It \new{employs} the Skip-Gram model to learn the dense vector representation for each byte and \new{adopts} the k-Nearest Neighbor (kNN) \new{algorithm} for classification. However, since the feature extraction and the classification are separated, the overall performance is limited by the individual optimization of each module. In contrast to Byte2Vec, Mittal~\textit{et al.} proposed an end-to-end method called FiFTy~\cite{mittal2020fifty} that can jointly optimize feature extraction and classification by using a 1D CNN. 
Saaim~\textit{et al.}~\cite{saaim2022light} further proposed a lightweight convolution network called DSCNN that utilizes 1D depthwise separable convolution as a replacement for the standard 1D convolution in FiFTy. This modification achieves faster inference time with comparable accuracy to FiFTy. In addition, Zhu~\textit{et al.}~\cite{zhu2023file} proposed a novel network \new{that merged a} short-term memory network (LSTM) for file analysis. While the CNN concentrates on learning high-level feature representations from file fragments, the LSTM specializes in classifying the fragments based on the learned features. 

There are relatively few studies that explore multimedia file fragment classification through visual perspectives. Chen~\textit{et al.}~\cite{chen2018file} were among the first to utilize CNNs for MFFC by reshaping 4096-byte sectors into 64×64 grayscale images by treating each byte as a pixel. Wang~\textit{et al.}~\cite{wang2023image} followed a similar approach for file type and malware classification, reshaping 512-byte memory blocks into 64×64 2D binary images by treating each bit as a pixel.

\subsection{Multimedia Analytics through Visual Perspectives}
With the development of CNNs, researchers are also exploring innovative approaches~\cite{bayoudh2021survey} from a visual perspective beyond traditional sequential processing for multimedia signal analysis, as vision signals are increasingly common~\cite{wang2024cm2, wu2022focal} and vision models are widely studied.
Visual representation stands as a good practice for audio signal recognition and analysis~\cite{lin2016audio, birajdar2020speech}. For instance, by utilizing a 2D representation of electric network frequency (ENF) signals, Lin~\textit{et al.} achieved superior performance and training efficiency compared to traditional audio recapture detection systems~\cite{lin2016audio}.
Birajdar~\textit{et al.}~\cite{birajdar2020speech} extracted chromatogram visual representations and uniform local binary pattern features from audio files, achieving good performance in speech and music signal classification. 
Extending beyond audio, researchers also utilize visual representations of other 1D signals, such as radio and EEG signals~\cite{wu2022rfmask, li2022unsupervised, yu2023mobirfpose, jiao2019decoding, song2022eeg}, to \new{develop} more effective signal analysis methods.
In~\cite{wu2022rfmask}, radio frequency (RF) signals are transformed into AoA-ToF (Angle of Arrival, Time of Flight) heatmaps, allowing effective segmentation of human silhouettes from millimeter-wave RF signals. Yu~\textit{et al.} transformed RF signals to RF heatmaps and utilized the generated heatmaps for 3D human pose estimation~\cite{yu2023mobirfpose}. Concurrently, Jiao~\textit{et al.}~\cite{jiao2019decoding} explored EEG maps as a visual representation for cognitive feature extraction, bridging the gap between cognitive and visual domains to improve \new{the categorization of EEG recording.} Song~\textit{et al.}~\cite{song2022eeg} further introduced the EEG conformer, \new{which treats} EEG signals as a visual representation akin to images for analysis, significantly improving the decoding quality of EEG signals.

Among the aforementioned studies using visual representations, the majority employ CNN-based vision models for feature extraction.  With the development of the vision model~\cite{liu2023bitstream}, self-attention mechanisms~\cite{vaswani2017attention, dosovitskiy2020image} are incorporated to improve the processing of input visual representations~\cite{song2022eeg}, significantly enhancing the models' capacity to capture long-term relationships within the data.

\begin{figure*}[t]
\centering
\includegraphics[width=7.0in]{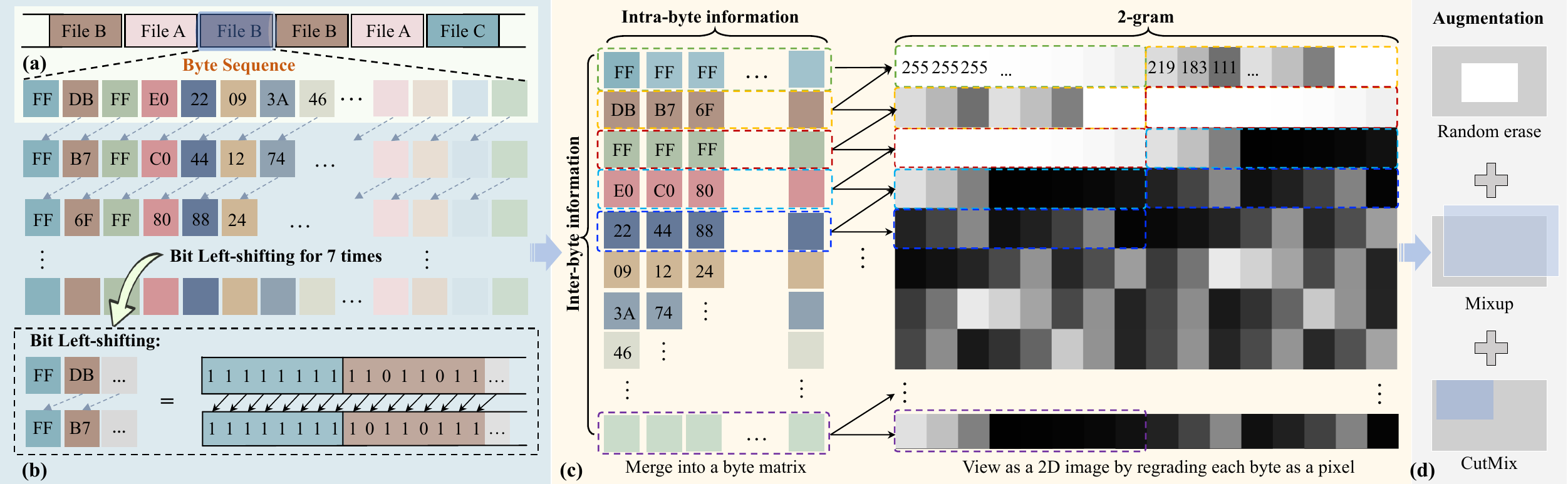} 
\caption{\textcolor{black}{Architecture} of the proposed \textbf{Byte2Image} representation model. (a) \textbf{Inputs} are fixed-sized memory sectors of different file types. (b) \textbf{\textcolor{black}{Intrabyte} information exposure and \textcolor{black}{interbyte} sequence stacking:} the overlooked \textcolor{black}{intrabyte} details within memory sectors are revealed by performing bit-shifting operations on the raw byte sequence seven times. Each bit-shifting operation exposes a different view of the byte's information. (c) \textbf{\textcolor{black}{Intrabyte} n-grams:} the \textcolor{black}{intrabyte} information acquired through bit-shifting operations is organized row by row, shaping a 2D byte matrix. We introduce \textcolor{black}{intrabyte} n-grams to increase the aspect ratio of this matrix and view it as a gray-scale image by regarding each byte as a pixel. (d) \textbf{Image augmentation:} the image undergoes augmentation using techniques like Random Erase and CutMix to diversify the training data.}
\label{f:byte2image}
\vspace{-0.15in}
\end{figure*}

\section{Methodology}
\subsection{Problem Statement}

The objective of multimedia file fragment classification is to establish a function $F$, which can accurately predict the class label $\hat{y}$ given the file fragment $x$:

\vspace{-0.1in}
\begin{equation}
 \hat{y} = F_\theta(x) 
\end{equation}

\noindent where $\theta$ represents the parameters of the function, $x$ represents a memory sector of fixed-size $N_s$ bytes, typically 512 or 4,096. 
% Recent advancements in machine learning have popularized learning-based approaches~\cite{mittal2020fifty, haque2022byte, saaim2022light} such as SVM, kNNs, and CNNs for modeling $F$, demonstrating superior performance.
For learning-based approaches, parameters $\theta$ for the function $F$ is determined by using a training dataset $\mathcal{D}$ consisting of $N$ sample pairs $\{(x^{(1)}, y^{(1)}), \dots, (x^{(N)}, y^{(N)})\}$. 
Once trained, the function $F$ is deployed to predict file types for unseen data samples.

Researchers are dedicated to designing various structures for modeling the function $F$, however, the overall performance is subjected to the limited representation capability of 1D byte sequence. 
To overcome this limitation, we propose Byte2Image, a novel visual representation that incorporates previously overlooked \new{intrabyte} information into memory sectors and reinterprets them as 2D images. 
Moreover, we propose a specialized classification method called ByteNet, which jointly considers the original raw 1D byte sequence and the converted 2D image.

\subsection{Proposed Byte2Image Representation Model}
\new{An} overview of our proposed Byte2Image model is shown in Fig.~\ref{f:byte2image}, which consists of the following three key components for converting raw 1D byte sequences into the 2D vision domain by using \new{intrabyte} information clues.

\textbf{\textit{\new{Intrabyte} information exposure and \new{interbyte} sequence stacking.}} As mentioned before, the overlooked \new{intrabyte} information is an important clue for classifying file sectors, which is uncovered by employing a sliding byte window with a stride of 1 bit in our design. To speed up the process, the sliding byte window is implemented by bit-shifting the raw $N_s$-byte sequence $x$ of byte sectors seven times (see (b) in Fig.~\ref{f:byte2image}). Each bit-shifting operation results in a new representation of the byte sequence. To construct the 2D byte matrix $\mathbf{x_{in}}$, we stack these resulting byte sequences from the bit shifts row by row (see (b) in Fig.~\ref{f:byte2image}) and transpose them for better illustration (see (c) in Fig.~\ref{f:byte2image}). The overall transformation can be expressed as:

\vspace{-0.1in}
\begin{equation}
x \xrightarrow{Bit Shift} \mathbf{x_{in}} \in \mathbb{Z}_{255}^{N_s \times 8} \ \ \ 
\mathbf{x_{in}} = [x_0^\mathsf{T}, x_1^\mathsf{T}, ... , x_7^\mathsf{T}]   
\end{equation}

\noindent where $x_0 = x \in \mathbb{Z}_{255}^{N_s}$ and $x_i = x_{i-1} << 1$. In the converted byte matrix $\mathbf{x_{in}}$ shown in Fig.~\ref{f:byte2image} (b), each row is composed of eight bytes where adjacent bytes differ by a 1-bit shifting distance, effectively introducing the previously overlooked \textbf{\textit{\new{intrabyte} information}}. Simultaneously, each column of this matrix $\mathbf{x_{in}}$ comprised of $N_s$ bytes, where adjacent bytes differ by an 8-bit shifting difference, equivalent to a 1-byte separation, representing the \textbf{\textit{\new{interbyte} information}}. Consequently, both \new{intrabyte and interbyte} information are fully contained in the byte matrix.

\textbf{\textit{\new{Intrabyte} n-grams.}} An intuitive idea for converting the byte matrix $\mathbf{x_{in}}$ into an image is to treat each byte as a pixel. In this scheme, each pixel corresponds to an 8-bit integer value, precisely the size of one byte. Therefore, $\mathbf{x_{in}}$ can be visualized as a grayscale image with a resolution of $N_s \times 8$, making it suitable for feature extraction via CNNs. However, since most commonly used CNNs~\cite{ghosh2019reshaping}, e.g., Resnet~\cite{he2016deep} and Densenet~\cite{huang2017densely}, are designed for square images, the relatively small aspect (the ratio of width to height), $8/N_s$, is not ideal for these CNNs to extract features, making it challenging to effectively extract deep features. To address this issue, inspired by the n-gram features~\cite{kim-2014-convolutional,tripathy2016classification} used in Natural Language Processing (NLP) for text sentence modeling, we regard each row of $\mathbf{x_{in}}$, which represents the \new{intrabyte} information, as a unigram feature. By using \new{intrabyte} n-grams to expand the image width, such as \new{the} 2-grams shown in Fig~\ref{f:byte2image} (c), we construct a new grayscale image denoted as $\mathbf{x_{ngram}}$. The overall transformation can be expressed as:

\vspace{-0.1in}
\begin{equation}
    \mathbf{x_{in}} \xrightarrow{Ngram} \mathbf{x_{ngram}} \in \mathbb{Z}_{255}^{H \times W \times 1}
\end{equation}

\noindent where $H = N_s - n + 1$ and $W=8n$. There are two notable advantages to using \new{intrabyte} n-grams. First, the aspect ratio of the converted images is increased, which is beneficial for the CNNs employed to extract features. Second, adjacent bytes in \new{intrabyte} n-grams still differ by a 1-bit shifting distance, consistent with the original \new{intrabyte} information, thereby preserving more relative order information in the converted images, which is beneficial for classification tasks.

% complex inter-byte and \new{intrabytes} correlations effectively.

% based on the complexity of the problem
\textbf{\textit{Image augmentation.}} Utilizing the proposed Byte2Image representation model to convert the raw 1D byte sequences into 2D images enables us to treat the multimedia file fragment classification problem as a standard image classification task. 
However, \new{it is} important to acknowledge that the number of data samples in the training dataset remains constant, which could potentially limit the adequacy of training and exacerbate the risk of overfitting, especially when utilizing a very deep network for the image classification task. 
To mitigate these issues, we adopt \new{the} image augmentation shown in Fig.~\ref{f:byte2image} (d) to generate additional new data samples from the n-gram images, enhancing the diversity of the training data:

\vspace{-0.1in}
\begin{equation}
    \mathbf{x_{ngram}} \xrightarrow{Aug} \mathbf{x_{im}} \in \mathbb{R}_{255}^{H \times W \times 1}
\end{equation}

\noindent where $Aug$ incorporates the following five image augmentation techniques. Image normalization and horizontal flipping are commonly used methods in image-related tasks~\cite{dosovitskiy2020image}, where the former involves adjusting the image using its mean and standard deviation, and the latter flips the image horizontally with a given probability. Random erase~\cite{zhong2020random} randomly selects a \new{rectangular} region in an image and erases its pixels with random values. This approach is particularly effective as \new{intrabyte} n-grams duplicate pixels, preventing the model from memorizing the repeated pattern and encouraging focus on \new{interbyte} and \new{intrabyte} correlations. CutMix~\cite{yun2019cutmix} and Mixup~\cite{zhang2017mixup} are techniques designed to explicitly augment training datasets by creating new samples through combinations with existing ones. CutMix selects a patch from one image and replaces it with a corresponding patch from another image in the dataset, while Mixup linearly interpolates between two randomly chosen samples and their corresponding labels. The incorporation of CutMix and Mixup \new{effectively prevents} model overfitting when using a deep vision network for image branch feature extraction.

\begin{figure*}[t]
\centering
\includegraphics[width=7.0in]{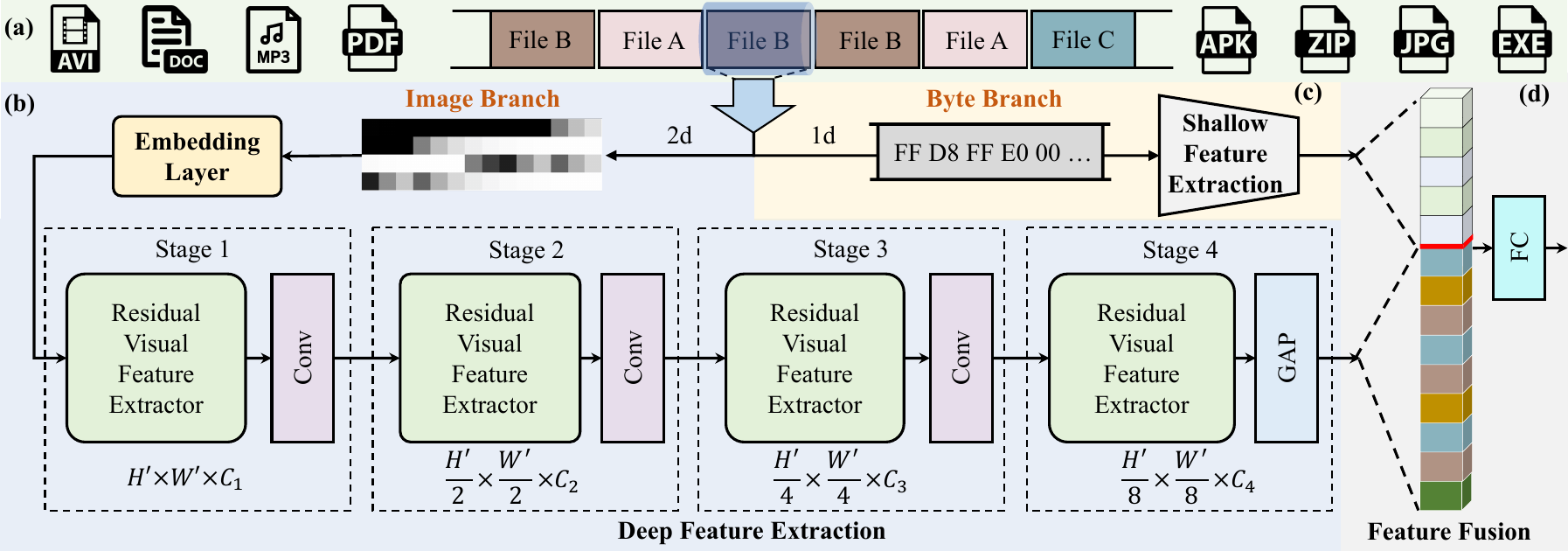} 
\caption{Architecture of the proposed \textbf{ByteNet} method, a dual-branch network designed to simultaneously handle raw 1D byte sequences and their corresponding 2D image representations. (a) \textbf{Inputs} are fixed-sized memory sectors of different file types. (b) \textbf{Image branch feature extraction} utilizes an embedding layer and a deep feature extraction module to extract complex features. The deep feature extraction module follows a hierarchical architecture of 4 stages, each comprising a residual visual feature extractor (RVFE) for feature extraction and a convolutional layer for downsampling features. The final stage incorporates a global average pooling (GAP) layer to flatten the feature maps. (c) \textbf{Byte branch feature extraction} employs a shallow feature extraction to capture the frequent co-occurrence of specific bytes. (d) \textbf{Feature fusion} is to concatenate extracted features from both the raw byte sequence and the image for the final classification.}
\label{f:Arch}
\vspace{-0.15in}
\end{figure*}

\subsection{Proposed ByteNet Classification Network} 

To make use of the information from the 1D raw byte sector and the converted 2D image, we propose an end-to-end dual-branch network ByteNet, \new{which consists} of a shallow byte branch feature extraction (BBFE) network and a deep image branch feature extraction (IBFE) network, \new{as} illustrated in Fig.~\ref{f:Arch}.

\textbf{\textit{Byte branch feature extraction.}} Magic bytes of file fragments can serve as valuable indicators for file fragment classification in certain file types, e.g., utilizing \textit{FF D8} bytes for classifying JPEG files, as demonstrated in earlier works~\cite{pal2009evolution}. To facilitate our model in memorizing these significant magic bytes, this process can be loosely defined as learning the frequent co-occurrence of specific bytes from the training datasets. Such co-occurrence of specific bytes can be regarded as shallow features, as they are directly observable from the raw byte sectors without requiring complex feature extraction. Hence, in our design shown in Fig~\ref{f:Arch} (c), we employ a shallow feature extraction component to capture these features. Specifically, we utilize a fully connected layer $H_{SF}(\cdot)$ to extract shallow features $\mathbf{x_{sf}}\in \mathbb{R}^{F_0}$ as follows

\vspace{-0.1in}
\begin{equation}
\mathbf{x_{sf}} = H_{SF}(x)
\end{equation} 

\noindent This module, utilizing a fully connected layer, provides a straightforward way to capture global features from byte sequences in contrast to the 1D CNN, demonstrating strong capabilities in globally memorizing the co-occurrence of specific bytes.

% \subsection{Byte2Image Branch Feature Extraction} 
\textbf{\textit{Image branch feature extraction.}} As illustrated in Fig.~\ref{f:Arch} (b), the module input $\mathbf{x_{im}}$ is 2D images with \new{intrabyte} n-grams and image augmentation techniques. Instead of handling the 2D images directly, we first employ a learnable embedding layer to enhance the fusion of n-gram information explicitly. The embedding layer has exhibited effectiveness in early-stage nature language processing~\cite{kim-2014-convolutional,tripathy2016classification} and visual processing in image Transformer~\cite{dosovitskiy2020image, liu2021swin}, leading to more stable optimization and increased flexibility. 
Specifically, the embedding layer projects discrete 2D images $\mathbf{x_{im}} \in \mathbb{R}^{H \times W \times 1}$ into dense 2D representations $\mathbf{x_{0}} \in \mathbb{R}^{H' \times W' \times C_1}$ as:

\vspace{-0.1in}
\begin{equation}
    \mathbf{x_{0}} = H_{EM}(\mathbf{x_{im}})
\end{equation}

\noindent Details of the embedding layer \new{are} given in the subsequent section. 

Then, we extract complex features through the proposed deep feature extraction module adopting a hierarchical architecture of 4 stages. Each stage $i$ contains a residual visual feature extractor (RVFE) and a $3 \times 3$ convolution layer for downsampling the scale of feature maps by half except for the last stage with a global average pooling (GAP) layer to flatten the feature maps. Each RVFE comprises $L_i$ feature extraction blocks in stage $i$. Specifically, intermediate feature maps $\mathbf{x_1}, \mathbf{x_2}, \mathbf{x_3}$ and the output deep feature $\mathbf{x_{df}}$ can be expressed as:

\vspace{-0.15in}
\begin{gather}
    \mathbf{x_i} = H_{CONV}(H_{RVFE_i}(\mathbf{x_{i-1}})), \ \  i=1, 2, 3 \\
    \mathbf{x_{df}} = H_{GAP}(H_{RVFE_i}(\mathbf{x_{i-1}}), \ \  i=4
\end{gather}

\noindent where $H_{RVFE_i}(\cdot)$ denotes the residual visual feature extractor employed in the $i$-stage, $H_{CONV}$ denotes the \new{following} convolutional layer and $H_{GAP}$ is the last GAP layer. 

In our RVFE module, we utilize two kinds of basic feature extraction blocks, i.e., \new{the} ResNet block~\cite{kalchbrenner2014convolutional} and \new{the} PoolFormer block~\cite{yu2023metaformer}. Within the ByteNet architecture, our initial design, named \textbf{ByteResNet}, employs the ResNet block, aligning with a classic CNN-based structure in RVFE modules. This design has exhibited performance enhancements in our previous research~\cite{liu2023byte}. Our alternative design, termed \textbf{ByteFormer}, integrates the PoolFormer block, inspired by recent advancements in parameter-efficient Transformers~\cite{dosovitskiy2020image, liu2021swin, yu2023metaformer}. This alteration enhances \new{the ability of the RVFE module} to capture long-term dependencies while drastically reducing the number of learned parameters.

\new{It is} worth noting that in the PoolFormer block, the token mixer is implemented by using a simple $pooling$ layer with stride 1, rather than a computationally intensive self-attention module. As a result, \new{there is} no explicit need to reshape the feature map $\mathbf{x_i}$ of the stage $i$ into a sequence of flattened 2D patched tokens. The feature map in each stage $i$ remains processed as a 2D image $\mathbf{x_i} \in \mathbb{R}^{ \frac{H'}{2^i} \times \frac{W'}{2^i} \times C_i}$, aligning with overall architecture in Fig~\ref{f:Arch} (b). Further details of these basic blocks will be given in the subsequent section.

\textbf{\textit{Feature fusion.}}
To fuse the extracted features from the raw byte sector and the converted image for classification, we \new{obtain} the final features by concatenating the corresponding shallow and deep features. These fused features are then utilized to obtain \new{the} predicted probabilities for each class, expressed as:

\vspace{-0.1in}
\begin{equation}
\hat{y} = \textit{Softmax}(H_{FC}(\mathbf{x_{sf}}, \mathbf{x_{df}}))
\end{equation}

\noindent where $H_{FC}$ represents the last fully connected layer with the output size equal to the number of classes, and \textit{Softmax} is the final activation layer used to normalize the outputs into probabilities.

\begin{figure}[t]
\centering
\includegraphics[width=3.5in]{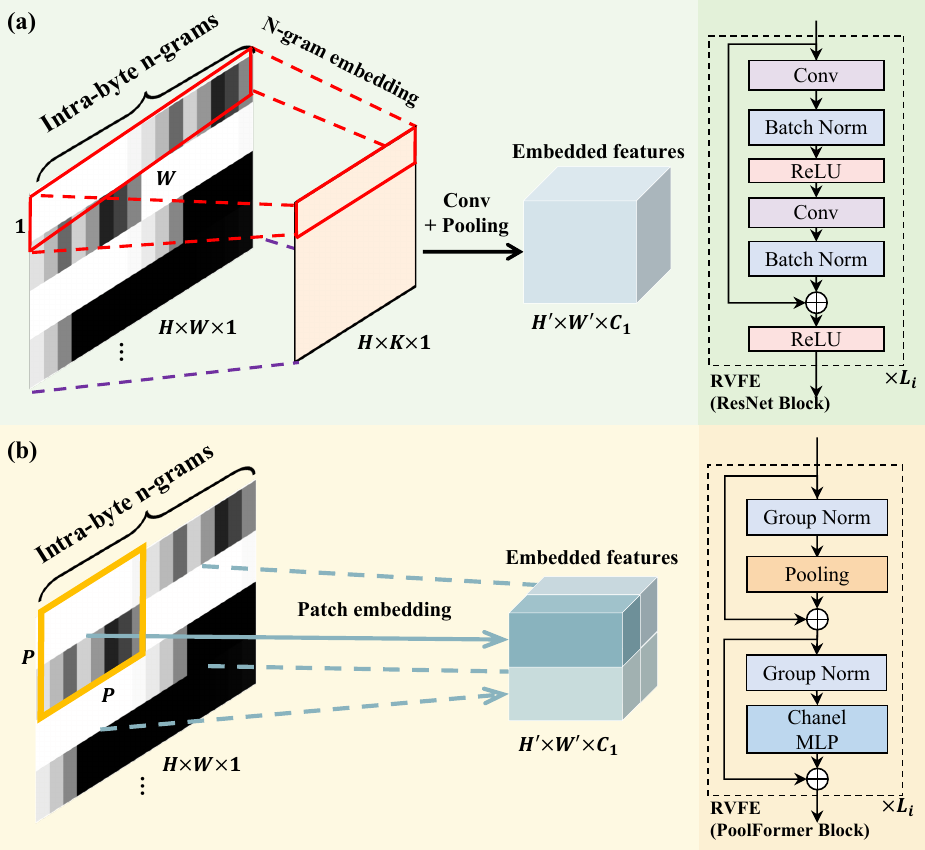}  
\caption{Details of two kinds of feature embedding layers and feature extraction blocks in the residual visual feature extractor (RVFE) include: (a) n-gram embedding with ResNet Block in \textbf{ByteResNet} and (b) patch embedding with PoolFormer block in \textbf{ByteFormer}.
}
\label{f:emb}
\vspace{-0.15in}
\end{figure}

\subsection{Embedding Layers and Feature Extraction Blocks} 
Within the ByteNet architecture, \new{which leverages} either CNNs or Transformers in the IBFE, ByteNet has two types of variants: ByteResNet and ByteFormer. These variants incorporate different types of embedding layers and feature extraction blocks, as depicted in Fig.~\ref{f:emb}.

\textbf{\textit{N-gram embedding layer in ByteResNet.}} In ByteResNet's embedding layer, we utilize the n-gram embedding layer (see Fig.~\ref{f:emb} (a)) to explicitly fuse the \new{intrabyte} n-grams information into intermediate features. This process involves a wide convolution (\new{the} kernel width is equal to the image width) to embed each row of sparse \textit{\new{intrabyte} n-grams} \new{into} a dense row vector by channel \new{concatenation}. Specifically, a wide convolution operation applies a weight matrix $W_i \in \mathbb{R}^{1 \times W \times 1}$ over the image $\mathbf{x_{im}} \in \mathbb{R}^{H \times W \times 1}$ to produce a 1D embedding feature $\mathbf{x_{emb, i}} \in \mathbb{R}^{H \times 1}$. A total of $K$ weight matrices are applied to \new{obtain} $K$ embeddings. These embeddings are then concatenated to \new{obtain} a 2D feature map $\mathbf{x_{emb}} \in \mathbb{R}^{H \times K \times 1}$, followed by a convolution layer with learnable parameters $W_0 \in \mathbb{R}^{C_1 \times 7 \times 7 \times 1}$ and a pooling layer to extract features $\mathbf{x_0}$ for the deep feature extraction module, expressed as: 

\vspace{-0.15in}
\begin{gather}
    \mathbf{x_{emb, i}} = W_i \mathbf{x_{im}}  \\
    \mathbf{x_{emb}} = [\mathbf{x_{emb, 1}}; \mathbf{x_{emb, 2}}; ...; \mathbf{x_{emb, K}}] \\
    \mathbf{x_0} = Pooling(W_0\mathbf{x_{emb}})
\end{gather}

\textbf{\textit{ResNet block in ByteResNet.}} In the deep feature extraction module, ByteResNet incorporates multiple ResNet blocks within each RVFE module. The ResNet block is based on the design introduced in~\cite{he2016deep}, which is a traditional CNN-based residual block. As depicted in Fig.~\ref{f:emb} (a), it comprises two convolutional layers with batch normalization and incorporates a shortcut connection. The input is passed through two convolutional layers, and the resulting output is added to the original input, forming the residual connection, expressed as:

\vspace{-0.1in}
\begin{equation}
    \mathbf{y} = \sigma(Norm(W_2\sigma(Norm(W_1\mathbf{x}))) + \mathbf{x})
\end{equation}

\noindent where $\mathbf{x}$ and $\mathbf{y}$ are \new{the} input and output features, \new{respectively}; $Norm(\cdot)$ is the batch normalization; $\sigma(\cdot)$ is a non-linear activation, such as ReLU~\cite{nair2010rectified} and GELU~\cite{hendrycks2016gaussian}; \new{and} $W_1 \in \mathbb{R}^{C \times 3 \times 3 \times C}$ and $W_2 \in \mathbb{R}^{C \times 3 \times 3 \times C}$ are \new{the} learnable parameters of the $3\times3$ convolution layers. 

\textbf{\textit{Patch embedding layer in ByteFormer.}} In ByteFormer's embedding layer, we follow the standard Transformer design by employing a patch embedding layer. \new{In contrast to} the n-gram embedding layer, the patch embedding layer treats the image $\mathbf{x_{im}}$ as a sequence of $P \times P$ 2D patches $\mathbf{x_p} \in \mathbb{R}^{N \times P^2}$ and maps each patch to \new{a} 1D embedding feature $\mathbf{x_{emb, i}}$ with a trainable linear projection $W_e \in \mathbb{R}^{P^2 \times C_1}$. As \textit{\new{intrabyte} n-grams} contains replicated \new{intrabyte} data from other rows, the patch embedding layer can also enable good \new{intrabyte} information fusion. These embeddings are then concatenated to form a new 2D feature map $\mathbf{x_{emb}} \in \mathbb{R}^{H' \times W' \times C_1}$. 
\new{The following output} is augmented with the position embedding $\mathbf{E_{pos}}$ to obtain $\mathbf{x_{0}}$, which is subsequently forwarded to the following deep feature extraction module for further feature extraction, expressed as:

\vspace{-0.15in}
\begin{gather}
    \mathbf{x_{emb, i}} = W_{e} \mathbf{x_{p}}  \\
    \mathbf{x_{emb}} = [\mathbf{x_{emb, 1}}; \mathbf{x_{emb, 2}}; ...; \mathbf{x_{emb, P^2}}] \\
    \mathbf{x_{0}} =  \mathbf{x_{emb}} + \mathbf{E_{pos}}
\end{gather}

\textbf{\textit{PoolFormer block in ByteFormer.}} In the deep feature extraction module, ByteFormer incorporates multiple PoolFormer blocks within each RVFE module. The PoolFormer block is based on a Transformer-based design introduced in~\cite{yu2023metaformer}. In conventional Transformer designs, a 2D image is typically divided into a sequence of image patches, and the long-term relations among them are extracted using attention-based token mixers. However, the attention-based token mixer is often criticized for its computational demands, specifically its quadratic time complexity concerning sequence length. To address this, PoolFormer blocks in ByteFormer replace the complex attention-based token mixer with a simple, non-parametric operator, a \textit{Pooling} layer. Consequently, there is no need to explicitly divide the feature map into a sequence of image patches. Specifically, as depicted in Fig.~\ref{f:emb} (b), PoolFormer blocks comprise two residual sub-blocks. The first sub-block employs a \textit{Pooling} layer as a token mixer, enabling efficient communication of information among tokens. The second sub-block consists of two layered channel MLP followed by non-linear activation, aiming to strengthen the model's representations. As the input is still a 2D image, the two-layered channel MLP is replaced by two $1\times1$ convolution layers. Mathematically, it can be expressed as:

\vspace{-0.15in}
\begin{gather}
    \mathbf{z} = Pooling(Norm(\mathbf{x})) + x \\
    \mathbf{y} = W_2\sigma(W_1(Norm(\mathbf{z}))) + \mathbf{z}
\end{gather}

\noindent where $\mathbf{x}$, $\mathbf{z}$ and $\mathbf{y}$ are \new{the} input, intermediate and output features, \new{respectively}; \new{and} $W_1 \in \mathbb{R}^{C \times 1 \times 1 \times rC}$ and $W_2 \in \mathbb{R}^{rC \times 1 \times 1 \times C}$ are learnable parameters with \new{an} MLP expansion ratio $r$.

\subsection{Training Loss}
The negative log-likelihood loss $\mathcal{L}_{NLL}$ is employed in both ByteResNet and ByteFormer to penalize the predicted label versus $\hat{y}$ the actual label $y$:

\vspace{-0.1in}
\begin{equation}
\label{e:theta}
\mathcal{L}_{NLL} (\hat{y}, y) = \sum_{n=1}^{N}  \sum_{t=1}^{T} -y^{(n)}_t {\rm log} \ \hat{y}^{(n)}_t
\end{equation}
\vspace{-0.1in}

\noindent where $\hat{y}^{(n)}_t$ and $y^{(n)}_t$ correspond \new{to} the predicted and actual probability of the $n$-th memory sector \new{of} file type $t$. As we utilize the Mixup augmentation technique to diversify the training samples, the actual label $y$ is transformed into a soft label based on the combination ratio between two samples belonging to different classes. The parameters of models are finally learned by minimizing the negative log-likelihood loss.

\section{Experiments}

\subsection{Dataset}
\textbf{\textit{FFT-75 dataset.}} FFT-75~\cite{mittal2020fifty} is the largest corpus for multimedia file fragment classification to date. It comprises 75 different file types. Each file type contains 102,400 sectors, with sector sizes of 512 and 4,096 bytes. \new{In particular,} 75 file types are categorized into 11 ``grouping tags'' such as Archive, Audio, etc., as indicated in Tab.~\ref{T:FFT}. The dataset defines 6 scenarios according to specific use cases, each scenario with different classification granularity. Specifically, \#1: it is the most challenging scenario required to classify all 75 file types into separate classes, whereas other scenarios are subsets following specific specifications; \#2: file types are classified into 11 grouped tags; \#3: except for 24 file types tagged with Bitmap, RAW, and Video are separate classes, all the remaining types are grouped as \textit{Others} class. \#4: file types such as JPEG and others tagged with Bitmap, RAW, and Video are viewed as 4 classes, and all the remaining types are grouped as the \textit{Others} class. \#5: JPEG vs \textit{Others}, where the remaining 74 file types are grouped as the \textit{Others} class. \#6: JPEG vs \textit{Others}, where RAW and 6 other video types (3gp, mov, mkv, tiff, and heic) are grouped as the \textit{Others} class. Tab~\ref{T:FFTVFF} details the number of classes and samples across these different scenarios.

\textbf{\textit{VFF-16 dataset.}} VFF-16~\cite{wang2024intra}~\footnote{https://ieee-dataport.org/documents/variable-length-file-fragment-dataset-vff-16; https://github.com/WangyiNTU/JSANet.} is a publicly available dataset derived from the GovDocs~\cite{stoykova2020standard}, a collection of 1 million documents from public sources. Specifically, it comprises 16 file types collected from the GovDocs, with each file type containing 50MB. These file types include jpg, gif, doc, xls, ppt, html, text, pdf, rtf, png, log, csv, gz, swf, eps, and ps. The dataset is partitioned into training and test sets \new{in an approximate ratio of 4:1.}
Tab~\ref{T:FFTVFF} details the number of classes and samples in this dataset.
We utilize the VFF-16 dataset to further evaluate our proposed methods' performance against other works. However, since some studies are not open-source, we provide evaluation results for some selected representative works.

\begin{table}[t]
\centering
\caption{Grouping tags of different file types on FFT-75.}
\resizebox{\linewidth}{!}{
% \begin{tabular}{p{1.8cm} p{3.9cm} l}
\begin{tabular}{l p{3.9cm} c }
\toprule
Grouping Tag (11) & Filetype (75) & Count \\  \midrule
Archive & apk, jar, msi, dmg, 7z, bz2, deb, gz,
pkg, rar, rpm, xz, zip & 13 \\  
Audio & aiff, flac, m4a, mp3, ogg, wav, wma & 7\\ 
Bitmap & jpg, tiff, heic, bmp, gif, png & 6\\ 
Executable & exe, mach-o, elf, dll & 4\\ 
Human-readable & md, rtf, txt, tex, json, html, xml, log, csv & 9\\ 
Office & doc, docx, key, ppt, pptx, xls, xlsx & 7\\ 
Published & djvu, epub, mobi, pdf  & 4\\ 
Raw & arw, cr2, dng, gpr, nef, nrw, orf, pef, raf, rw2, 3fr & 11 \\ 
Vector & ai, eps, psd  & 3\\ 
Video & mov, mp4, 3gp, avi, mkv, ogv, webm & 7\\ 
Miscellaneous & pcap, ttf, dwg, sqlite & 4\\ 
\bottomrule
% \hline Total & FFT-75 dataset & 75 \\
%  11 Tags &  & 75 \\ \hline
\end{tabular}
}
\label{T:FFT}
\vspace{-0.15in}
\end{table}

% \subsection{Experimental settings}
\subsection{Implementation Details}
We adopt the AdamW~\cite{loshchilov2017decoupled} optimizer with $(\beta_1, \beta_2)=(0.9, 0.999)$ and \new{a weight decay of 0.01.} The training batch sizes in FFT-75 and VFF-16 are set to 512 and 256, respectively. The learning rate is warmed up from $5\text{e-}7$ to $5\text{e-}4$ linearly in the first 2 epochs and then decays to 0 via the cosine scheduler in the \new{remaining} 48 epochs. We implement our model with the PyTorch on two NVIDIA GeForce RTX 3090 GPUs. The sector size $N_s$ is set to 512 or 4,096. The n-gram $n$ is set to 16. For image augmentation techniques, we set the probability parameter $\alpha$ for \new{the} employed Normalization, Horizontal flip, Random erase, CutMix, and Mixup to 1.0, 0.5, 0.5, 1.0, and 0.8 respectively. 

The proposed ByteNet has two architecture settings: ByteResNet and ByteFormer. For ByteResNet, the channel dimension of the n-gram embedding layer is set to 96 and the number of ResNet blocks $L_i$ in each stage is set to [2, 2, 2, 2], and the corresponding channel dimension $C_i$ in each stage is set to [64, 128, 256, 512]. For ByteFormer, the channels dimension of the patch embedding layer is set to 64 and 96 for \new{the} 512-byte sectors and \new{the} 4,096-byte sectors, respectively and the patch size is set to 8. The number of PoolFormer blocks $L_i$ in each stage is set to [6, 6, 18, 6], and the corresponding channel dimension $C_i$ in each stage is set to [64, 128, 320, 512]. It should be noted that the simplicity of the non-parametric $Pooling$ token mixer employed in ByteFormer allows for potentially much deeper architectures than ByteResNet under similar memory and computational resource constraints. 
% in FFT-75 and VFF-16 datasets.

One issue we encounter when processing 4,096-byte sectors is the considerable computational complexity arising due to the converted image, \new{which reaches} a 4k resolution with the height $H=N_s-n+1, N_s=4096$. While employing resizing operations or downsampling modules could alleviate this computational load, we find that both methods incur significant information loss and subsequent performance degradation. Given our model's strong performance in 512-byte sectors, we aim to adopt this model to address the \new{challenges} posed by 4,096-byte sectors. Specifically, our strategy involves segmenting the 4,094-byte memory sector into eight 512-byte parts. Each part undergoes conversion into a grayscale image using the proposed Byte2Image technique. These eight grayscale images are then concatenated across the channel and fed into our models. This strategy effectively reduces the computational load while preserving the same processing step as our model in processing 512-byte sectors, ensuring a seamless adaptation to larger sector sizes.

\begin{table}[t]
  \caption{Specifications of different scenarios on FFT-75 and VFF-16.}
   \centering
    \begin{tabular}{l l c c}
\toprule 
Dataset & Scenarios & \#Classes  & \#Samples \\ \midrule \multirow{6}{*}{FFT-75}  & \#1 (512 \& 4,096) & 75 &  7500k \\ 
& \#2 (512 \& 4,096) & 11 &  1935k \\ 
& \#3 (512 \& 4,096) & 25 &  2300k  \\ 
& \#4 (512 \& 4,096) &  5 &  1054k  \\ 
& \#5 (512 \& 4,096) &  2 & 1036k   \\ 
& \#6 (512 \& 4,096) & 2  & 1000k   \\ \hline

\multirow{2}{*}{VFF-16} &  \#1 (512)   & 16  &  1639k  \\ 
 &   \#1 (4,096)   & 16  &  209k \\ 

\bottomrule
\end{tabular}

    \label{T:FFTVFF}
    \vspace{-0.15in}
\end{table}

\begin{table*}[]
\centering
\caption{Comparing state-of-the-art methods in scenario \#1 of FFT-75 with a sector size (SS) of 512 and 4,096 bytes. Sceadan, NN-GF, and NN-CO employ hand-crafted features, while Byte2Vec, FiFTy, DSCNN, DSCNN-SE, and CNN+LTSM use 1D byte-embedded features from raw byte sequences. DIA utilizes constructed 2D binary images by reshaping byte sequences into a 2D space.}
\vspace{-0.05in}
% The best and second-best performances are highlighted in \textcolor{red}{red} and \textcolor{blue}{blue}, respectively. 
\label{T:scen1}
\resizebox{\linewidth}{!}{

\belowrulesep=0pt
\aboverulesep=0pt

\begin{tabular}{| l | l | c | c | c c c  | c c | c c |} 
% \multirow{2}{*}{Method}  & \multirow{2}{*}{Hand} & \multirow{2}{*}{1d}  & \multirow{2}{*}{2d}  & \multicolumn{2}{c}{SS = 512} \\ \cline{5-6} 
\hline
\multirow{2}{*}{Method} & \multirow{2}{*}{Year} &  \multirow{2}{*}{Stage} & \multirow{2}{*}{Architecture} & \multicolumn{3}{c|}{Input Features}  & \multicolumn{2}{c|}{\#1 (SS = 512)} & \multicolumn{2}{c|}{\#1 (SS = 4,096)} \\ \cline{5-11}
  & & & & Crafted & 1D & 2D  & Acc. & JPEG Acc.  & Acc. & JPEG Acc. \\ \hline
 Sceadan~\cite{beebe2013sceadan} & 2013  & Two & SVM  & \cmark  &  \xmark &  \xmark & 57.3 & 81.5 &  69.0 & 91.8 \\ 
NN-GF~\cite{mittal2020fifty} & 2019  & Two &  NN  & \cmark  &  \xmark &  \xmark  &  45.4 & 53.2 & 46.8 & 48.6  \\ 
NN-CO~\cite{mittal2020fifty}  & 2019   & Two &  2D CNN  & \cmark  &  \xmark &  \xmark  &  64.4 & 77.0 & 75.3 & 92.5 \\   

Byte2Vec+kNN~\cite{haque2022byte} & 2022  & Two &  KNN  &  \xmark & \cmark & \xmark  &   50.1 & 42.4 & - & -\\   
FiFTy~\cite{mittal2020fifty} & 2020 &  One & 1D CNN  &  \xmark & \cmark & \xmark  &  65.6 & 83.5 & 77.5 & 86.3 \\
DSCNN~\cite{saaim2022light}  & 2022 & One  & 1D Depth-Separable CNN  &  \xmark & \cmark & \xmark  & 65.9 & 83.4 & 78.5 & 91.7 \\
DSCNN-SE~\cite{ghaleb2023file} & 2023  & One   & 1D DSCNN + SE module  & \xmark & \cmark & \xmark  & 66.3 & - & 79.3 & - \\ 
CNN-LTSM~\cite{zhu2023file} &  2023 & One  &  1D CNN + LTSM module  &  \xmark & \cmark & \xmark  & 66.5 & 87.0 & 78.6 & 90.3 \\
 % \hdashline
DIA~\cite{wang2023image} &  2023 & One & 1D inception-attention network &  \xmark & \xmark  & \cmark   & 68.3 &  87.5 & 81.0 & 90.6 \\ 

\hdashline
 
ByteResNet (Ours) & 2023  & One &  2D ResNet block   &  \xmark & \cmark & \cmark & 71.0 & 89.2 & \textbf{82.1} & 93.4 \\ 
ByteFormer (Ours) & 2023  & One  & 2D PoolFormer block  &  \xmark  & \cmark & \cmark & \textbf{73.2} & \textbf{96.0} & 81.9 & \textbf{95.1} \\ \hline
\end{tabular}

}
\vspace{-0.1in}
\end{table*}

\begin{table*}[h]    
\centering
\caption{Comparison with state-of-the-art methods in other scenarios from \#2 to \#6 of FFT-75 with a sector size (SS) of 512 and 4,096 bytes, where the number of classes of each scenario required to be classified is given. “N.A.” means there is no JPEG class in this scenario.}
% The best and second best performances are in \textcolor{red}{red} and \textcolor{blue}{blue} colors, respectively.
\vspace{-0.05in}
\label{T:scen2}
\resizebox{\linewidth}{!}{

\belowrulesep=0pt
\aboverulesep=0pt

\begin{tabular}{|l | c | cc | cc | cc| cc | cc |} 
\hline
\multirow{2}{*}{Method}  & \multirow{2}{*}{SS} &\multicolumn{2}{c|}{\#2 (11 classes)}   & \multicolumn{2}{c|}{\#3 (25 classes)}    & \multicolumn{2}{c|}{\#4 (5 classes)}  &  \multicolumn{2}{c|}{\#5 (2 classes)}   & \multicolumn{2}{c|}{\#6 (2 classes)}  \\ \cmidrule{3-12}
        &     & Acc. & JPEG Acc. & Acc. & JPEG Acc. & Acc. & JPEG Acc. & Acc. & JPEG Acc. & Acc. & JPEG Acc.  \\ \hline 

FiFTy~\cite{mittal2020fifty}    & 512     & 78.9 & N.A. &  87.9 & 93.3 & 90.2 & \textbf{98.6} &  99.0 & 99.3  &  99.3 &  99.5  \\ 

DSCNN~\cite{saaim2022light} & 512  &  75.8 &  N.A. &  80.8  & 94.6 & 87.1 & 97.3 & 98.9 & 98.9 & 98.8 & 98.6 \\

DSCNN-SE~\cite{ghaleb2023file} & 512  &  75.0 &  N.A. &  80.8  & - & 87.3 & - & 99.0 & - & 98.7 & - \\  

DIA~\cite{wang2023image} & 512 & 89.5 & N.A. & 92.8 & - & 92.8 & - & 99.3 & - & 99.2 &  -  \\ 
\hdashline
ByteResNet (Ours) &   512   &  90.4 & N.A. & 93.5 & 96.6 & 93.6 & 97.4 & 99.2 & 99.3  & 99.2 & 99.2 \\ 
ByteFormer (Ours) & 512  & \textbf{91.1}   & N.A. & \textbf{93.9} & \textbf{97.8} & \textbf{94.0}  & 98.4 & \textbf{99.5} & \textbf{99.6} & \textbf{99.5} & \textbf{99.6}  \\  \hline 

FiFTy~\cite{mittal2020fifty}    & 4,096  &  89.8  & N.A. & 94.6 & \textbf{98.9} &  94.1 & \textbf{99.1} & 99.2 & 99.2 & 99.6 & \textbf{99.7}   \\ 

DSCNN~\cite{saaim2022light} & 4,096  & 85.7 & N.A. & 93.1 & 96.8 & 94.2 & 98.4 & 99.3 & 99.3 & 99.6 & \textbf{99.7} \\

DSCNN-SE~\cite{ghaleb2023file} & 4,096  &  87.1 &  N.A. &  93.3  & - & 94.6 & - & \textbf{99.4} & - & \textbf{99.7} & \textbf{99.7}\\  
DIA~\cite{wang2023image} & 4,096 &  94.0 & N.A. &  96.7 & - & \textbf{96.1} & - & 99.2 & - & 99.5 & - \\  
\hdashline
ByteResNet (Ours) & 4,096  & \textbf{94.2}   & N.A. & \textbf{96.8} & 98.1 & \textbf{96.1}  & 98.6 & 99.3 & \textbf{99.6} & 99.4  & 99.5   \\ 
ByteFormer (Ours) & 4,096  &  93.8  & N.A. & 96.7 & 98.6 & 95.7  & 98.3 & 99.1 & 99.4 & 99.3 &  99.5  \\

\hline
\end{tabular}

}
\vspace{-0.15in}
\end{table*}

\subsection{Experimental Setting}
\textbf{\textit{Evaluation metrics.}} Following~\cite{mittal2020fifty}, we report the average classification accuracy for all file types on \new{the} testing datasets.
Additionally, the classification accuracy of JPEG files is also provided following~\cite{mittal2020fifty}. 

% given that JPEG files are one of the most commonly used file types.

\textbf{\textit{Baseline methods.}} We compare them against the following state-of-the-art baseline methods, which are categorized into two groups.

\begin{figure*}[t]
\centering
\includegraphics[width=7.0in]{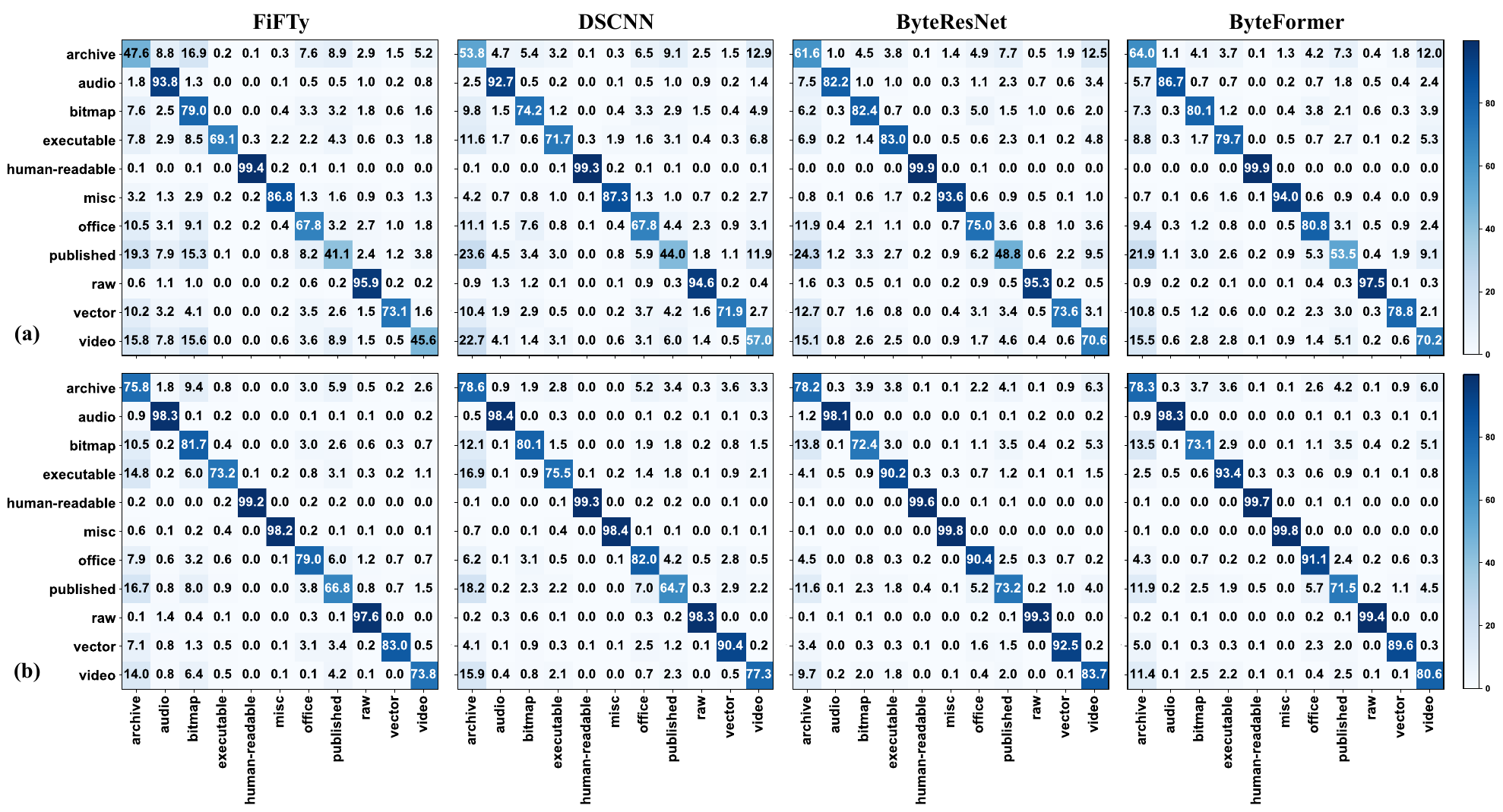} 
\vspace{-0.1in}
\caption{\textcolor{black}{Comparison of the confusion matrices of} FiFTy, DSCNN, and our proposed methods, ByteResNet and ByteFormer, evaluating in scenario \#1 of FFT-75. Due to the large number of classes (75), classes belonging to the same superclass were clustered into one. (a) SS = 512. (b) SS = 4,096. }
\label{f:CM}
\vspace{-0.1in}
\end{figure*}

\begin{figure*}[t]
\centering
\includegraphics[width=6.4in]{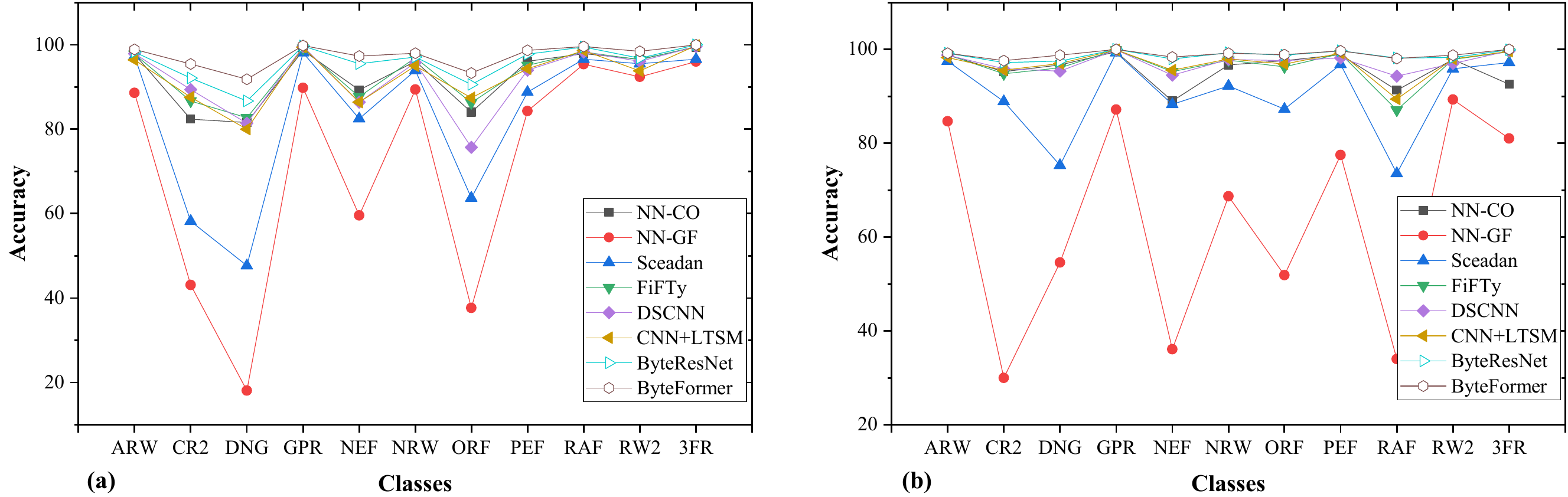} 
\vspace{-0.1in}
\caption{Accuracy curves for each file type within the grouped 'Raw' tags evaluated in scenario \#1 on the FFT-75. The accuracy fluctuations reflect the models' sensitivity within use-case tags. (a) SS = 512. (b) SS = 4,096. }
\label{f:DS}
\vspace{-0.25in}
\end{figure*}

1) Early works primarily rely on \new{handcrafted} features for classification, where the quality of the selected features often \new{affects the performance more than the employed classifier does}: Sceadan~\cite{beebe2013sceadan} utilizes statistical features \new{such as} unigram or entropy features and employs Support Vector Machines (SVM) for classification; NN-GF~\cite{mittal2020fifty} uses a fully-connected neural network trained with 14 specifically selected global statistical features for classification; NN-CO~\cite{mittal2020fifty} uses a shallow 2D convolutional neural network with 6 layers trained on specifically defined byte co-occurrence features. 

2) Recent studies \new{have focused} on learning features from byte sequences for classification. These studies can be categorized \new{as two-stage or one-stage}. Two-stage approaches, \new{such as} Byte2Vec+kNN~\cite{haque2022byte}, separate feature extraction and classification for individual learning, \new{resulting in a decrease in performance.} In contrast, one-stage approaches, exemplified by FiFTy, focus on concurrently extracting and classifying features from byte sequences, often demonstrating improved performance. Other one-stage methodologies, including DSCNN~\cite{saaim2022light}, DSCNN-SE~\cite{ghaleb2023file}, and CNN-LTSM~\cite{zhu2023file}, build upon FiFTy's core concept of using inter-byte embeddings as features with slight modifications to the classifier's architecture. For instance, \new{the} DSCNN incorporates \new{a} 1D depthwise separable CNN instead of \new{a} 1D CNN, and \new{the} DSCNN+SE integrates a Squeeze-and-Excitation block, resulting in limited performance improvements.

\subsection{Experimental Results on the FFT-75 Dataset}
\textbf{\textit{Comparison of accuracy on the most challenging scenario.}} Table~\ref{T:scen1} shows \new{a} comparison of results \new{for} \#1 of the FFT-75 dataset \new{for sector sizes (SSs)} of 512 and 4,096 bytes, which is the most complicated scenario than others. Notably, all models deliver better results at SS = 4,096 than \new{at} SS = 512, as the larger sector size inherently contains more information and richer byte relationships. One-stage approaches exhibit superior performance compared to two-stage approaches. 
Compared with these one-stage approaches, our proposed method, ByteResNet, outperforms FiFTy and DIA by 5.4\% and 4.6\%, \new{respectively}, achieving the best results with the average accuracy of 71.0\% and 82.1\% in sector sizes of 512 and 4,096 bytes. \new{These} impressive results emphasize the superiority of our proposed ByteResNet, \new{which is} empowered by the Byte2Image representation model. 
\new{In addition}, benefiting from the Transformer design and employed PoolFormer blocks, ByteFormer shows an additional gain of 2.2\% and comparable performance to ByteResNet, achieving average accuracies of 73.2\% and 81.9\% \new{for} sector sizes of 512 and 4,096 bytes, respectively.
Given the large scale of FFT-75, the improvement of our methods is significant, e.g., a 1\% improvement means an increase of 7,680 correct predictions over 768,000 test samples. 
\new{For} the JPEG (using variable length coding) accuracy \new{for} sector sizes of 512 and 4,096 bytes, our proposed method, ByteFormer, \new{exhibits notably superior performance compared with other approaches}, showing the best results with JPEG \new{accuracies} of 96.0\% and 95.1\%, respectively. These impressive results underline the superiority of our proposed method in uncovering overlooked \new{intrabyte} information by looking inside the byte.

\textbf{\textit{Comparison of accuracy \new{in} the remaining scenarios.}} Table~\ref{T:scen2} shows the comparison of results in the remaining scenarios of the FFT-75 dataset in sector size (SS) of 512 and 4,096 bytes. In these scenarios, our proposed methods, ByteResNet and ByteFormer, outperform one-stage approaches \new{such as} FiFTy, DSCNN, DSCNN-SE, and DIA significantly. Notably, ByteResNet and ByteFormer exhibit comparable performances, with ByteFormer tending to achieve superior results in SS = 512 and ByteResNet tending to excel in SS = 4,096. The improvements are particularly evident in complex scenarios \#2 and \#3. For instance, compared with the FiFTy baseline, our method ByteResNet improves the average accuracy by 11.5\% and 5.6\% in SS = 512 \new{at} scenarios \#2 and \#3, \new{respectively}, achieving the best results with the \new{accuracies} of 90.4\% and 93.5\%, respectively. \new{In addition}, ByteFormer show an additional gain of 0.4\% and 1\% \new{at} SS = 512 compared to ByteResNet. As the number of classes required to be classified decreases in scenarios \#4, \#5 and \#6, the classification task becomes relatively easier, enabling all methods to perform well. For example, in the 2-class classification tasks (JPEG-vs-others) of \#5 and \#6 scenarios, all methods show comparable performance, achieving nearly 99\% accuracy.

\textbf{\textit{Comparison of accuracy on the grouping tags.}}
To show the comparison results more intuitively, we give the \new{normalized} confusion matrices of some baseline works (FiFTy and DSCNN) and our proposed methods, (ByteResNet and ByteFormer) \new{for \#1 of the FFT-75 dataset} for both sector sizes of 512 and 4,096 bytes in Fig.~\ref{f:CM}. In these normalized confusion tables, the file types are grouped \new{into} 11 superclasses based on \new{the} use-case tags in Tab.~\ref{T:FFT} for \new{clarity}. 
Notably, most misclassifications among these models occur within the 'Archive' and 'Published' file groups.
This tendency arises due to the potential embedding of other file types \new{such as} Bitmap and Audio, within these categories, leading to misleading classifications.
Our proposed method, ByteFormer, shows the most promising results, \new{as} indicated by the dense diagonal units in Fig.~\ref{f:CM}, \new{showing} high accuracy rates for each use-case tag. 
Notably, it exhibits significant improvements, particularly in the 'Video' tag compared to other models. 
To further evaluate model sensitivity within use-case tags, we compare the accuracy curves for each file type within the grouped 'Raw' tags at SS = 512 and SS = 4,096, as illustrated in Fig.~\ref{f:DS}. Our models demonstrate superior performance with smooth accuracy fluctuations. 
We believe that models showing smooth accuracy fluctuations within use-case tags can effectively learn fundamental discriminating features, given that file types within a use-case tag often exhibit higher similarities than other file types.

\subsection{Experimental Results on the VFF-16 Dataset}
\begin{figure}[t]
  \centering
  \includegraphics[width=3.1in]{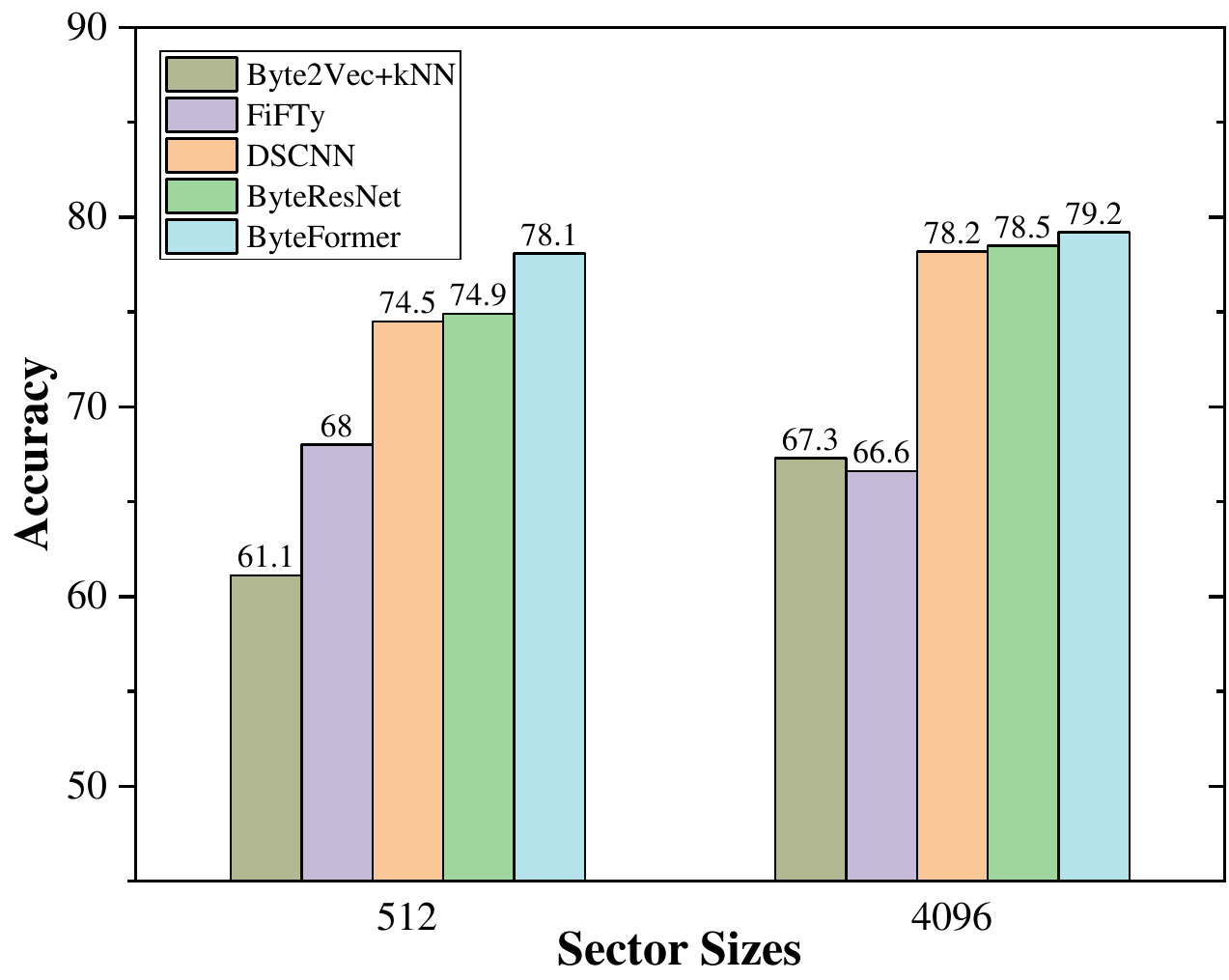}
    \vspace{-0.1in}
  \caption{Comparison with the baseline methods including FiFTy, Byte2Vec+kNN, DSCNN, and our proposed methods, ByteResNet and ByteFormer, in the VFF-16 test dataset.}
  \label{fig:VFF16}
  \vspace{-0.2in}
\end{figure}

\textbf{\textit{Comparison of accuracy.}} Fig~\ref{fig:VFF16} shows \new{a} comparison of results \new{for} \#1 of the VFF-16 dataset for sector sizes (SSs) of 512 and 4,096 bytes.
At SS = 512, one-stage methods such as FiFTy and DSCNN demonstrate superior performance over the two-stage method Byte2Vec+kNN. 
At SS = 4,096, FiFTy marginally underperforms compared to Byte2Vec+kNN, causing a slight 0.7\% decrease in overall accuracy, while DSCNN achieves a significant 10.9\% improvement over Byte2Vec+kNN. Our proposed model, ByteResNet, achieves 74.9\% accuracy at SS = 512 and 78.5\% at SS = 4,096, surpassing all state-of-the-art methods. Furthermore, ByteFormer shows an additional gain of 3.2\% and 0.7\% compared to ByteResNet, outperforming all methods, especially at SS = 512, achieving up to 78.1\%. 
Interestingly, while classifying byte sectors with longer lengths is assumed to be easier than \new{classifying byte sectors with shorter lengths because they contain more information,} our proposed model, ByteFormer, shows limited improvement at SS = 4,096 compared to \new{that at} SS = 512. Other baseline models, \new{such as FiFTy}, even experience an accuracy decrease. This discrepancy might be attributed to the significantly smaller number of training samples in SS = 4,096 as depicted in Table~\ref{T:FFTVFF}, comprising only 1/8 of the samples at SS = 512. Consequently, ByteFormer might struggle due to the reduced training samples at SS = 4,096, hindering its complete training potential. In addition, we observed that decreasing the batch size from 256 to 16 in the VFF-16 training dataset at SS = 4,096 resulted in improved accuracy for ByteFormer. This observation suggests that the reduced training samples at SS = 4,096 could help achieve ByteFormer's full training potential, subsequently improving the model's performance.

\begin{figure*}[t]
\centering
\includegraphics[width=7.1in]{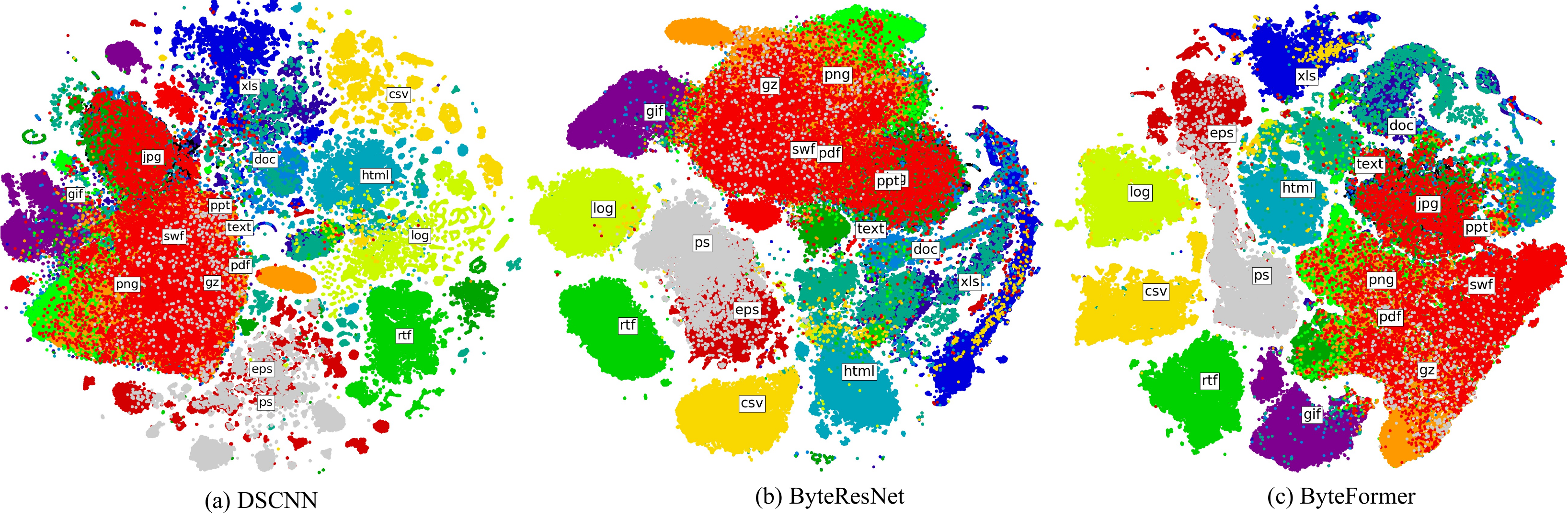} 
\vspace{-0.35in}
\caption{Visualization of feature representations extracted by \textcolor{black}{the DSCNN baseline method} and our proposed methods, ByteResNet and ByteFormer, in the VFF-16 testing dataset using t-SNE~\cite{van2008visualizing} for dimensionality reduction. The text label is located in the centroids of each file type cluster. }
\label{f:tsne}
\vspace{-0.15in}
\end{figure*}

\textbf{\textit{Comparison of feature representation capability.}} The capability of a robust feature representation significantly enhances the classifier's decision-making process, enabling clear differentiation of feature descriptions within the embedding space. We believe that a good model should possess this capability to establish a robust feature representation before the decision-making process. To evaluate this, we employ t-distributed stochastic neighbor embedding (t-SNE)~\cite{van2008visualizing} to reduce the dimensionality of feature representations to 2, allowing visualization in a 2D space. \new{In particular}, t-SNE employs a non-linear unsupervised technique for dimensionality reduction, offering valuable insights into the distribution of high-dimensional features. We select the state-of-the-art DSCNN method as the baseline for comparison against our proposed methods. Fig.~\ref{f:tsne} presents the visualization of feature representations for \new{all the} samples in the testing dataset of VFF-16 at SS = 512. Notably, DSCNN fails to cluster data samples of file types such as ps, log, rtf, html, and csv into distinct clusters. Conversely, our proposed models, ByteResNet and ByteFormer, exhibit superior performance, where data samples for these file types form distinct clusters, enabling easier separation. Compared with the feature representation of ByteResNet, in ByteFormer, file types \new{such as} xls can be clustered into a tight group and the centroids of file types \new{such as} png and pdf move farther away, demonstrating a more robust feature representation.

\vspace{-0.1in}
\subsection{Discussion}
% \textbf{\textit{Discussion.}} 
Existing methods in MFFT, e.g., FiFTy, DSCNN, and DSCNN-SE are notable \new{because of their ability to customize} multiple network architectures to optimize performance in different scenarios. However, these methods often involve large tuning of hyper-parameters based on Tree-structured Parzen Estimator~\cite{hutter2011sequential} algorithm, including layer numbers and kernel sizes, to adapt to diverse scenarios. For instance, FiFTy employs 12 different network architectures to test 6 scenarios with 2 different sector sizes in test datasets. Each architecture is selected from a total of 225 models to obtain the best candidate for the respective scenario. While this approach can enhance performance for specific scenarios, it severely constrains the flexibility of the models in handling other datasets and scenarios. 

One possible explanation for FiFTy's heavy reliance on customizing multiple network architectures extensively is the susceptibility to underfitting due to the shallowness of the employed network. This issue is similarly noted in Mittal's study~\cite{mittal2020fifty}, where the employed FiFTy model achieves comparable performance on the validation dataset even with less than 20\% of the training data. This phenomenon suggests that the model may encounter difficulties in capturing essential deep patterns, limiting its ability to fully leverage the extensive training dataset. 

In contrast, our proposed methods, ByteResNet and ByteFormer, benefit from deeper architectures by treating the MFFT problem as an image classification task. 
It enables our proposed models to employ the same architecture to adapt to all scenarios without suffering any underfitting issues. Consequently, our models exhibit greater adaptability and flexibility. 

\begin{table}[t]
\centering
\label{T:ex}
\caption{Ablation study on the effectiveness of the image branch and byte branch, which is evaluated in scenario \#2 of the FFT-75 dataset (SS = 512) and scenario \#1 of the VFF-16 dataset (SS = 512).}
  \vspace{-0.05in}
% \resizebox{\linewidth}{!}{

\begin{tabular}{| l |  cc |}
\hline
\multirow{2}{*}{Method} &  \multicolumn{2}{c|}{Accuracy (\%)} \\ 
&  FFT-75 (\# 2) & VFF-16 (\# 1) \\ \hline
ByteResNet (byte branch only)  &  30.0  & 27.5 \\ 
ByteResNet (image branch only)   &  93.5   & 74.5  \\ 
ByteResNet (Full) &  \textbf{93.6}  & \textbf{74.9}  \\  \hline
ByteFormer (byte branch only)  &  30.0  & 27.5 \\ 
ByteFormer (image branch only) & 93.9 & 77.4 \\ 
ByteFormer (Full) &   \textbf{94.0}   & \textbf{78.1} \\ \hline
\end{tabular}

% }
\label{t:byte_image}
  \vspace{-0.1in}
\end{table}

\vspace{-0.1in}
\subsection{Ablation Study}
\textbf{\textit{Effectiveness of byte and image branch.}} To further validate the effectiveness of the dual-branch design in our proposed methods, ByteResNet and ByteFormer, we conduct an ablation study on the byte and image branches. This comparison is performed in scenario \#2 of the FFT-75 dataset \new{with the} sector size of 512 and in scenario \#1 of the VFF-16 dataset \new{with the} sector size = 512. The results are detailed in Table~\ref{t:byte_image}. Notably, when removing the image branch from our proposed models, ByteResNet and ByteFormer, the performance significantly decreases to less than 30\% \new{for} both FFT-75 and VFF-16. \new{Moreover}, when removing the byte branch, the decrease in accuracy is relatively marginal, decreasing by only 0.1\% and less than 1\% \new{on} FFT-75 and VFF-16 datasets, respectively. These results suggest that both the image and byte branches contribute independently to \new{improving} accuracy, \new{demonstrating} their effectiveness.

\begin{table}[]
\centering
\label{T:ex}
\caption{Model complexity for different N-gram settings, which is evaluated in scenario \#4 of the FFT-75 dataset (SS = 512).}
\vspace{-0.05in}
\begin{tabular}{| c |  c | c c c |}
\hline
Method & N-gram &  \#Params. (M) & \#MACs (G) & Acc. (\%) \\ \hline
\multirow{4}{*}{ByteResNet} & 2 & 11.24 & 1.71 & 90.4  \\
 & 4 & 11.24 & 1.71 & 90.4 \\
 & 8 & 11.25 & 1.70 & 90.8 \\
 & 16 & 11.25 & 1.67 & 89.8 \\ \hline

\multirow{4}{*}{ByteFormer} & 2 & 30.35 & 0.41 & 89.8  \\
 & 4 & 30.35 & 0.46 & 90.4 \\
 & 8 & 30.35 & 0.81 & 90.8 \\
 & 16 & 30.35 & 1.61 & 91.1 \\ \hline
\end{tabular}
\label{fig:patch1}
\vspace{-0.1in}
\end{table}

\begin{table}[t]
\centering
\label{T:ex}
\caption{Ablation study on the patch size of ByteFormer, which is evaluated in scenario \#2 of the FFT-75 dataset (SS = 512).}
\vspace{-0.05in}
\begin{tabular}{| c |  c | c c c |}
\hline
Method & Patch Size & \#Param. (M) & \#MACs (G) & Acc. (\%) \\ \hline
\multirow{4}{*}{ByteFormer} & (8, 8) & 30.36 & 1.61 & 91.1  \\
& (16, 16) & 30.37 & 0.41 & 90.2 \\
& (24, 24) & 30.39 & 0.24 & 89.2 \\
& (32, 32) & 30.42 & 0.12 & 89.0 \\ \hline

\end{tabular}
\label{fig:patch}
\vspace{-0.2in}
\end{table}

\begin{table}[t]
\centering
\label{T:ex}
\caption{Model complexity comparison of FiFTy, DSCNN, ByteResNet, and ByteFormer, which is evaluated in scenario \#2 of the FFT-75 dataset (SS = 512).}
\vspace{-0.05in}
\begin{tabular}{| c |  c c c |}
\hline
Method & \#Param. (M) & \#MACs (G) & Acc. (\%) \\ \hline
FiFTy &  0.27 & 0.07 & 78.9  \\
DSCNN &  0.09 & 0.01 & 75.8\\ \hline
ByteResNet & 11.25 & 1.67 & 90.4 \\
ByteFormer & 30.35 & 1.61 & 91.1\\ \hline
ByteFormer-S & 30.42 & 0.12 & 89.0 \\ \hline
\end{tabular}
\label{fig:baseline}
% \vspace{-0.2in}
\end{table}

\begin{figure}[t]
  \centering
  \includegraphics[width=2.8in]{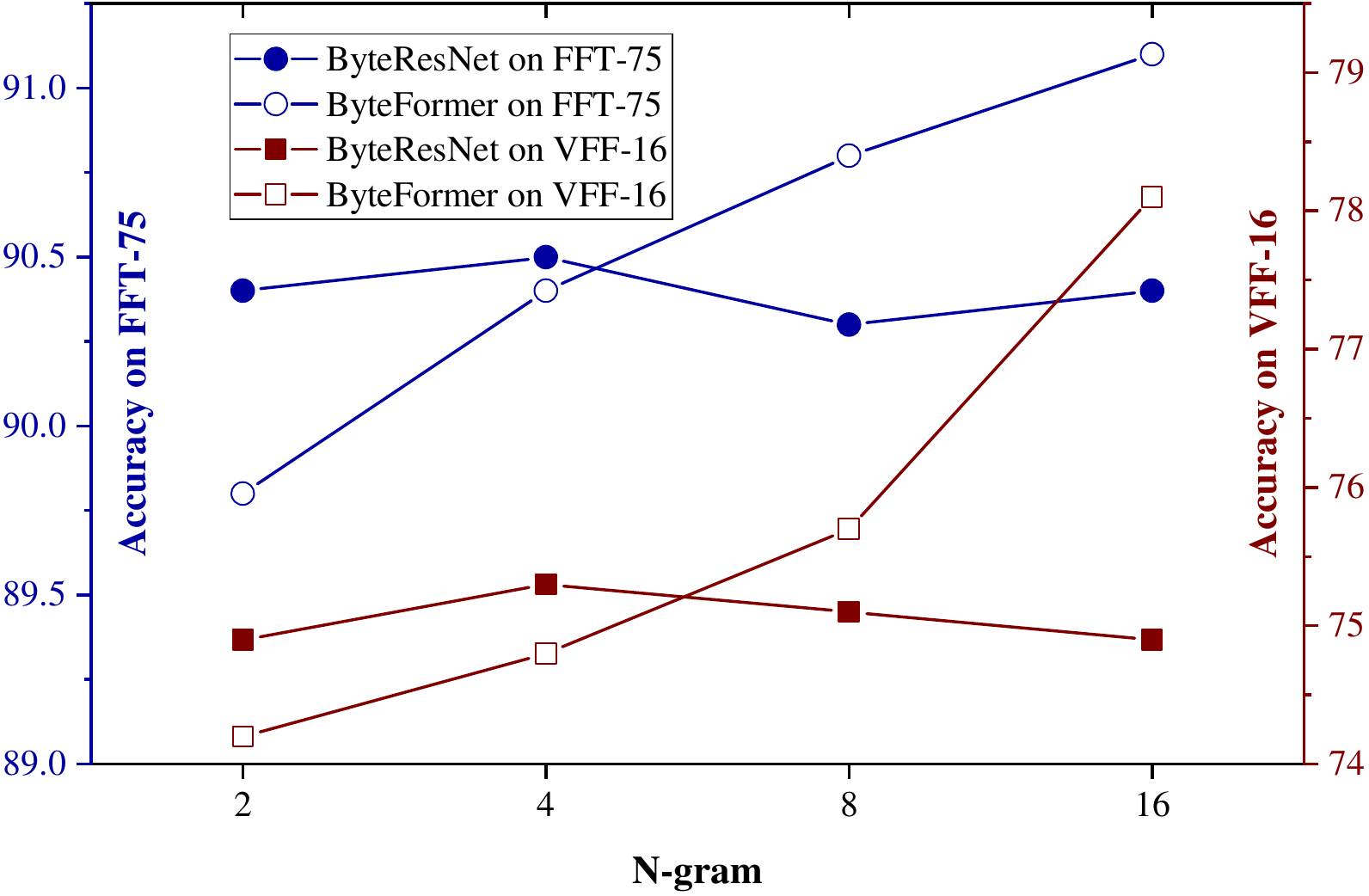}
    \vspace{-0.1in}
  \caption{Ablation study on the effectiveness of the different settings of the n-gram set, which is evaluated in scenario \#4 of the FFT-75 dataset (SS = 512) and the scenario \#1 of the VFF-16 dataset (SS = 512).}
  \label{fig:n-gram}
  \vspace{-0.2in}
\end{figure}

\textbf{\textit{Effectiveness of different N-gram settings.}} We also evaluate the impact of different settings for the n-gram set. This comparison was conducted in scenario \#4 of the FFT-75 dataset \new{with a} sector size of 512 and in scenario \#1 of the VFF-16 dataset \new{with a} sector size of 512. 
The results for various n-gram settings of $2^{1, 2, 3, 4} = {2, 4, 8, 16}$ are presented in Fig~\ref{fig:n-gram} and \new{the} corresponding computation costs are given in Table~\ref{fig:patch1}. 
\new{The} computational cost of ByteFormer is significantly reduced with a smaller n-gram setting, while ByteResNet is not reduced due to the fixed size \new{of} n-gram embedding. Notably, ByteFormer shows significantly improved performance with larger n-gram settings, reaching its peak performance in both datasets when the n-gram is set to 16. \new{An increase in the n-gram size allows} for more spatial information, leading to enhanced performance for ByteFormer. In contrast, ByteResNet appears to be less sensitive to the n-gram settings, displaying an oscillating trend of performance with fluctuating accuracies. This might be attributed to the introduced n-gram embedding layer in ByteResNet, \new{which enables} the model to fuse \new{intrabyte} n-gram information initially before being sent to the deep feature extraction module. On the other hand, the introduced patch-embedding layer in ByteFormer requires the model to fuse \new{intrabyte} n-gram information stage by stage due to the patch division. This process may encounter challenges with smaller n-gram settings due to the reduction in the number of tokens.

\textit{\textbf{Effectiveness of different patch size settings.}} We also evaluated the impact of patch size in ByteFormer. The results for various patch size settings ${8, 16, 24}$ \new{and} ${32}$, as well as the computation costs are presented in Table~\ref{fig:patch}. As observed, ByteFormer benefits from smaller patch sizes, achieving the best performance when the patch size is set to 8. This is because, given that the size of the converted images remains unchanged, smaller patch sizes allow more image tokens to be considered simultaneously, though slightly increasing the computational cost.

% Consequently, this enhances the model's capacity to capture fine-grained details and local features within the image for classification.

\textit{\textbf{Model complexity analysis.}} We conduct experiments to analyze the complexity of our method compared that of existing methods, i.e., FiFTy and DSCNN. As shown in Table~\ref{fig:baseline}, existing methods typically employ fewer parameters due to their use of shallow layers and 1D convolutions, which significantly limits their model performance. In contrast, our model, \new{which leverages} the proposed Byte2Image representation, can fully unleash the potential of larger parameters, enabling the capture of deep and complex features and ultimately achieving superior results. Furthermore, benefiting from the computation-efficient PoolFormer block design, our proposed ByteFormer allows for a deeper and more complex network, while preserving less computational cost compared to ByteResNet. 
Additionally, ByteFormer allows for more flexible consideration by utilizing a larger patch size, such as (8, 8), termed \textbf{ByteFormer-S}, which can achieve superior results with comparable computational costs.

\section{Future Work}
\subsection{Performance Improvement}
The proposed ByteNet architecture addresses the MFFC through visual perspectives. Out of necessity, we only explore several commonly used image classification network designs, which are primarily optimized for processing images. It is possible to develop better architectures tailored for MFFC. Additional improvements in classiﬁcation accuracy can be expected by considering certain characteristics inherent in multimedia file fragments~\cite{gibert2018classification}, e.g., high information entropy, multiple formats representing the same content, and \new{a} high aspect ratio when converted to images.

Due to the limitations of using standard benchmark datasets, we have only evaluated our proposed methods on unencrypted file types. However, since encryption does not alter the contents of files, it is possible that our model can learn relevant features from encrypted data without the need for a decryption pre-process, as long as we construct encrypted file-type datasets. 
In future work, we plan to explore commonly used file types that undergo malicious alteration including encryption, insertion, or corruption, and leverage our models to detect these malicious behaviors.

\subsection{Content-based Classification}
The multimedia file fragment classification tasks aim to extract essential features from byte-level raw data for file-type classification. In contrast to content-based classification tasks, such as image or text classification, our extracted features typically carry less semantic information. This can be attributed to two main factors. \new{First}, the training target in file-type classification is focused on identifying file formats, which are often unrelated to semantic information. \new{Second}, byte-level raw data convey less semantic information than normal images \new{because} they predominantly represent byte-level information and are subject to various file encoding methods. These factors make semantic information mining from byte-level data challenging in the context of byte-based file-type classification.

One approach to addressing this challenge could involve leveraging newly developed network architectures, such as Transformers and Large Language Models (LLMs)~\cite{zhao2023survey}, and changing the training paradigm, such as next byte prediction or next token prediction~\cite{wu2024beyond}. Recently, some novel works have shifted their focus towards proposing a universal framework that allows models to work with byte-level representations of text, images, and other types of data. For example, a recent study~\cite{horton2023bytes} based on Transformer directly handles raw byte sequences converted from images and audio to achieve image and audio classification. Other works, \new{such as} MambaByte~\cite{wang2024mambabyte} and bGPT~\cite{wu2024beyond} leverage the Mamba network and LLMs, exhibiting greater potential in exploring semantic information over byte-level representation. These advancements offer exciting opportunities for advancing the semantic understanding of byte-level representations of files.

\color{black}
\section{Conclusion}\label{sec13}
To achieve more effective multimedia file fragment classification (MFFC), this paper innovatively represents and analyzes the features of multimedia file fragments from a visual perspective.
\new{Specifically}, we first propose Byte2Image, a novel visual representation model, to introduce the overlooked \new{intrabyte} information into 1D byte sequences of file fragments and reinterpret them as 2D images. As the 1D raw byte sequence and the converted 2D image are two different representations of file fragments, we further propose an end-to-end dual-branch network ByteNet, \new{which incorporates} a shallow byte branch feature extraction network and a deep image branch feature extraction network, to enhance robust correlation mining and feature representation. Built on either CNNs or Transformers in the deep image branch feature extraction (IBFE), ByteNet has two types of variants called ByteResNet and ByteFormer, which can both capture the essential \new{interbyte} and \new{intrabyte} correlations for classification. Extensive experiments on benchmarks show our method greatly exceeds state-of-the-art methods on multimedia file type classification, which demonstrates the superiority of our proposed methods.

\section*{Acknowledgment}
This research is supported by the National Research Foundation, Singapore, and Cyber Security Agency of Singapore under its National Cybersecurity Research \& Development Programme (Cyber-Hardware Forensic \& Assurance Evaluation R\&D Programme$<$NRF2018NCRNCR009-0001$>$). Any opinions, findings and conclusions or recommendations expressed in this material are those of the author(s) and do not reflect the view of National Research Foundation, Singapore and Cyber Security Agency of Singapore

\bibliographystyle{ieeetr}

\bibliography{cas-refs}

\begin{thebibliography}{10}

\bibitem{sundararaj2021opposition}
V.~Sundararaj and M.~Selvi, ``Opposition grasshopper optimizer based multimedia data distribution using user evaluation strategy,'' {\em Multimedia Tools and Applications}, vol.~80, pp.~29875--29891, 2021.

\bibitem{cai2024towards}
C.~Cai, K.-H. Yap, and S.~Wang, ``Towards attribute-controlled fashion image captioning,'' {\em ACM Transactions on Multimedia Computing, Communications and Applications}, 2024.

\bibitem{cai2024top}
C.~Cai, S.~Wang, K.-H. Yap, and Y.~Wang, ``Top-down framework for weakly-supervised grounded image captioning,'' {\em Knowledge-Based Systems}, vol.~287, p.~111433, 2024.

\bibitem{deldjoo2020recommender}
Y.~Deldjoo, M.~Schedl, P.~Cremonesi, and G.~Pasi, ``Recommender systems leveraging multimedia content,'' {\em ACM Computing Surveys (CSUR)}, vol.~53, no.~5, pp.~1--38, 2020.

\bibitem{wu2024multifocal}
K.~Wu, Q.~Liu, K.-H. Yap, and Y.~Yang, ``Multifocal multiview imaging and data compression based on angular--focal--spatial representation,'' {\em Optics Letters}, vol.~49, no.~3, pp.~562--565, 2024.

\bibitem{ohm2015transmission}
J.-R. Ohm and J.~Ohm, ``Transmission and storage of multimedia data,'' {\em Multimedia Signal Coding and Transmission}, pp.~491--520, 2015.

\bibitem{liu2024bitstream}
T.~Liu, K.~Wu, Y.~Wang, W.~Liu, K.-H. Yap, and L.-P. Chau, ``Bitstream-corrupted video recovery: a novel benchmark dataset and method,'' {\em Advances in Neural Information Processing Systems}, vol.~36, 2024.

\bibitem{uzun2015carving}
E.~Uzun and H.~T. Sencar, ``Carving orphaned jpeg file fragments,'' {\em IEEE transactions on Information Forensics and Security}, vol.~10, no.~8, pp.~1549--1563, 2015.

\bibitem{pal2009evolution}
A.~Pal and N.~Memon, ``The evolution of file carving,'' {\em IEEE signal processing magazine}, vol.~26, no.~2, pp.~59--71, 2009.

\bibitem{beebe2013sceadan}
N.~L. Beebe, L.~A. Maddox, L.~Liu, and M.~Sun, ``Sceadan: using concatenated n-gram vectors for improved file and data type classification,'' {\em IEEE Transactions on Information Forensics and Security}, vol.~8, no.~9, pp.~1519--1530, 2013.

\bibitem{haque2022byte}
M.~E. Haque and M.~E. Tozal, ``Byte embeddings for file fragment classification,'' {\em Future Generation Computer Systems}, vol.~127, pp.~448--461, 2022.

\bibitem{mittal2020fifty}
G.~Mittal, P.~Korus, and N.~Memon, ``Fifty: large-scale file fragment type identification using convolutional neural networks,'' {\em IEEE Transactions on Information Forensics and Security}, vol.~16, pp.~28--41, 2020.

\bibitem{horton2023bytes}
M.~Horton, S.~Mehta, A.~Farhadi, and M.~Rastegari, ``Bytes are all you need: Transformers operating directly on file bytes,'' {\em arXiv preprint arXiv:2306.00238}, 2023.

\bibitem{he2016deep}
K.~He, X.~Zhang, S.~Ren, and J.~Sun, ``Deep residual learning for image recognition,'' in {\em Proceedings of the IEEE conference on computer vision and pattern recognition}, pp.~770--778, 2016.

\bibitem{dosovitskiy2020image}
A.~Dosovitskiy, L.~Beyer, A.~Kolesnikov, D.~Weissenborn, X.~Zhai, T.~Unterthiner, M.~Dehghani, M.~Minderer, G.~Heigold, S.~Gelly, {\em et~al.}, ``An image is worth 16x16 words: Transformers for image recognition at scale,'' {\em arXiv preprint arXiv:2010.11929}, 2020.

\bibitem{ghosh2019reshaping}
S.~Ghosh, N.~Das, and M.~Nasipuri, ``Reshaping inputs for convolutional neural network: Some common and uncommon methods,'' {\em Pattern Recognition}, vol.~93, pp.~79--94, 2019.

\bibitem{liu2023byte}
W.~Liu, Y.~Wang, K.~Wu, K.-H. Yap, and L.-P. Chau, ``A byte sequence is worth an image: Cnn for file fragment classification using bit shift and n-gram embeddings,'' in {\em 2023 IEEE 5th International Conference on Artificial Intelligence Circuits and Systems (AICAS)}, pp.~1--5, 2023.

\bibitem{mcdaniel2003content}
M.~McDaniel and M.~H. Heydari, ``Content based file type detection algorithms,'' in {\em 36th Annual Hawaii International Conference on System Sciences, 2003. Proceedings of the}, pp.~10--pp, IEEE, 2003.

\bibitem{li2005fileprints}
W.-J. Li, K.~Wang, S.~J. Stolfo, and B.~Herzog, ``Fileprints: Identifying file types by n-gram analysis,'' in {\em Proceedings from the Sixth Annual IEEE SMC Information Assurance Workshop}, pp.~64--71, IEEE, 2005.

\bibitem{karresand2006oscar}
M.~Karresand and N.~Shahmehri, ``Oscar—file type identification of binary data in disk clusters and ram pages,'' in {\em IFIP International Information Security Conference}, pp.~413--424, Springer, 2006.

\bibitem{wang2018sparse}
F.~Wang, T.-T. Quach, J.~Wheeler, J.~B. Aimone, and C.~D. James, ``Sparse coding for n-gram feature extraction and training for file fragment classification,'' {\em IEEE Transactions on Information Forensics and Security}, vol.~13, no.~10, pp.~2553--2562, 2018.

\bibitem{saaim2022light}
K.~M. Saaim, M.~Felemban, S.~Alsaleh, and A.~Almulhem, ``Light-weight file fragments classification using depthwise separable convolutions,'' in {\em IFIP International Conference on ICT Systems Security and Privacy Protection}, pp.~196--211, Springer, 2022.

\bibitem{zhu2023file}
N.~Zhu, Y.~Liu, K.~Wang, and C.~Ma, ``File fragment type identification based on cnn and lstm,'' in {\em Proceedings of the 2023 7th International Conference on Digital Signal Processing}, pp.~16--22, 2023.

\bibitem{chen2018file}
Q.~Chen, Q.~Liao, Z.~L. Jiang, J.~Fang, S.~Yiu, G.~Xi, R.~Li, Z.~Yi, X.~Wang, L.~C. Hui, {\em et~al.}, ``File fragment classification using grayscale image conversion and deep learning in digital forensics,'' in {\em 2018 IEEE Security and Privacy Workshops (SPW)}, pp.~140--147, IEEE, 2018.

\bibitem{wang2023image}
Y.~Wang, K.~Wu, W.~Liu, K.-H. Yap, and L.-P. Chau, ``Image representation and deep inception-attention for file-type and malware classification,'' in {\em 2023 IEEE International Symposium on Circuits and Systems (ISCAS)}, pp.~1--5, IEEE, 2023.

\bibitem{bayoudh2021survey}
K.~Bayoudh, R.~Knani, F.~Hamdaoui, and A.~Mtibaa, ``A survey on deep multimodal learning for computer vision: advances, trends, applications, and datasets,'' {\em The Visual Computer}, pp.~1--32, 2021.

\bibitem{wang2024cm2}
R.~Wang, C.~Cai, W.~Wang, J.~Gao, D.~Lin, W.~Liu, and K.-H. Yap, ``Cm2-net: Continual cross-modal mapping network for driver action recognition,'' {\em arXiv preprint arXiv:2406.11340}, 2024.

\bibitem{wu2022focal}
K.~Wu, Y.~Yang, Q.~Liu, and X.-P. Zhang, ``Focal stack image compression based on basis-quadtree representation,'' {\em IEEE Transactions on Multimedia}, vol.~25, pp.~3975--3988, 2022.

\bibitem{lin2016audio}
X.~Lin, J.~Liu, and X.~Kang, ``Audio recapture detection with convolutional neural networks,'' {\em IEEE Transactions on Multimedia}, vol.~18, no.~8, pp.~1480--1487, 2016.

\bibitem{birajdar2020speech}
G.~K. Birajdar and M.~D. Patil, ``Speech/music classification using visual and spectral chromagram features,'' {\em Journal of Ambient Intelligence and Humanized Computing}, vol.~11, pp.~329--347, 2020.

\bibitem{wu2022rfmask}
Z.~Wu, D.~Zhang, C.~Xie, C.~Yu, J.~Chen, Y.~Hu, and Y.~Chen, ``Rfmask: A simple baseline for human silhouette segmentation with radio signals,'' {\em IEEE Transactions on Multimedia}, 2022.

\bibitem{li2022unsupervised}
T.~Li, L.~Fan, Y.~Yuan, and D.~Katabi, ``Unsupervised learning for human sensing using radio signals,'' in {\em Proceedings of the IEEE/CVF Winter Conference on Applications of Computer Vision}, pp.~3288--3297, 2022.

\bibitem{yu2023mobirfpose}
C.~Yu, D.~Zhang, Z.~Wu, C.~Xie, Z.~Lu, Y.~Hu, and Y.~Chen, ``Mobirfpose: Portable rf-based 3d human pose camera,'' {\em IEEE Transactions on Multimedia}, 2023.

\bibitem{jiao2019decoding}
Z.~Jiao, H.~You, F.~Yang, X.~Li, H.~Zhang, and D.~Shen, ``Decoding eeg by visual-guided deep neural networks.,'' in {\em IJCAI}, pp.~1387--1393, Macao, 2019.

\bibitem{song2022eeg}
Y.~Song, Q.~Zheng, B.~Liu, and X.~Gao, ``Eeg conformer: Convolutional transformer for eeg decoding and visualization,'' {\em IEEE Transactions on Neural Systems and Rehabilitation Engineering}, vol.~31, pp.~710--719, 2022.

\bibitem{liu2023bitstream}
W.~Liu, Y.~Wang, K.-H. Yap, and L.-P. Chau, ``Bitstream-corrupted jpeg images are restorable: Two-stage compensation and alignment framework for image restoration,'' in {\em Proceedings of the IEEE/CVF Conference on Computer Vision and Pattern Recognition}, pp.~9979--9988, 2023.

\bibitem{vaswani2017attention}
A.~Vaswani, N.~Shazeer, N.~Parmar, J.~Uszkoreit, L.~Jones, A.~N. Gomez, {\L}.~Kaiser, and I.~Polosukhin, ``Attention is all you need,'' {\em Advances in neural information processing systems}, vol.~30, 2017.

\bibitem{huang2017densely}
G.~Huang, Z.~Liu, L.~Van Der~Maaten, and K.~Q. Weinberger, ``Densely connected convolutional networks,'' in {\em Proceedings of the IEEE conference on computer vision and pattern recognition}, pp.~4700--4708, 2017.

\bibitem{kim-2014-convolutional}
Y.~Kim, ``Convolutional neural networks for sentence classification,'' in {\em Proceedings of the 2014 Conference on Empirical Methods in Natural Language Processing ({EMNLP})}, (Doha, Qatar), pp.~1746--1751, Association for Computational Linguistics, Oct. 2014.

\bibitem{tripathy2016classification}
A.~Tripathy, A.~Agrawal, and S.~K. Rath, ``Classification of sentiment reviews using n-gram machine learning approach,'' {\em Expert Systems with Applications}, vol.~57, pp.~117--126, 2016.

\bibitem{zhong2020random}
Z.~Zhong, L.~Zheng, G.~Kang, S.~Li, and Y.~Yang, ``Random erasing data augmentation,'' in {\em Proceedings of the AAAI conference on artificial intelligence}, vol.~34, pp.~13001--13008, 2020.

\bibitem{yun2019cutmix}
S.~Yun, D.~Han, S.~J. Oh, S.~Chun, J.~Choe, and Y.~Yoo, ``Cutmix: Regularization strategy to train strong classifiers with localizable features,'' in {\em Proceedings of the IEEE/CVF international conference on computer vision}, pp.~6023--6032, 2019.

\bibitem{zhang2017mixup}
H.~Zhang, M.~Cisse, Y.~N. Dauphin, and D.~Lopez-Paz, ``mixup: Beyond empirical risk minimization,'' {\em arXiv preprint arXiv:1710.09412}, 2017.

\bibitem{liu2021swin}
Z.~Liu, Y.~Lin, Y.~Cao, H.~Hu, Y.~Wei, Z.~Zhang, S.~Lin, and B.~Guo, ``Swin transformer: Hierarchical vision transformer using shifted windows,'' in {\em Proceedings of the IEEE/CVF international conference on computer vision}, pp.~10012--10022, 2021.

\bibitem{kalchbrenner2014convolutional}
N.~Kalchbrenner, E.~Grefenstette, and P.~Blunsom, ``A convolutional neural network for modelling sentences,'' {\em arXiv preprint arXiv:1404.2188}, 2014.

\bibitem{yu2023metaformer}
W.~Yu, C.~Si, P.~Zhou, M.~Luo, Y.~Zhou, J.~Feng, S.~Yan, and X.~Wang, ``Metaformer baselines for vision,'' {\em IEEE Transactions on Pattern Analysis and Machine Intelligence}, 2023.

\bibitem{nair2010rectified}
V.~Nair and G.~E. Hinton, ``Rectified linear units improve restricted boltzmann machines,'' in {\em Proceedings of the 27th international conference on machine learning (ICML-10)}, pp.~807--814, 2010.

\bibitem{hendrycks2016gaussian}
D.~Hendrycks and K.~Gimpel, ``Gaussian error linear units (gelus),'' {\em arXiv preprint arXiv:1606.08415}, 2016.

\bibitem{wang2024intra}
Y.~Wang, W.~Liu, K.~Wu, K.-H. Yap, and L.-P. Chau, ``Intra-and inter-sector contextual information fusion with joint self-attention for file fragment classification,'' {\em Knowledge-Based Systems}, vol.~291, p.~111565, 2024.

\bibitem{stoykova2020standard}
R.~Stoykova and K.~Franke, ``Standard representation for digital forensic processing,'' in {\em 2020 13th International Conference on Systematic Approaches to Digital Forensic Engineering (SADFE)}, pp.~46--56, IEEE, 2020.

\bibitem{loshchilov2017decoupled}
I.~Loshchilov and F.~Hutter, ``Decoupled weight decay regularization,'' {\em arXiv preprint arXiv:1711.05101}, 2017.

\bibitem{ghaleb2023file}
M.~Ghaleb, K.~Saaim, M.~Felemban, S.~Al-Saleh, and A.~Al-Mulhem, ``File fragment classification using light-weight convolutional neural networks,'' {\em arXiv preprint arXiv:2305.00656}, 2023.

\bibitem{van2008visualizing}
L.~Van~der Maaten and G.~Hinton, ``Visualizing data using t-sne.,'' {\em Journal of machine learning research}, vol.~9, no.~11, 2008.

\bibitem{hutter2011sequential}
F.~Hutter, H.~H. Hoos, and K.~Leyton-Brown, ``Sequential model-based optimization for general algorithm configuration,'' in {\em Learning and Intelligent Optimization: 5th International Conference, LION 5, Rome, Italy, January 17-21, 2011. Selected Papers 5}, pp.~507--523, Springer, 2011.

\bibitem{gibert2018classification}
D.~Gibert, C.~Mateu, J.~Planes, and R.~Vicens, ``Classification of malware by using structural entropy on convolutional neural networks,'' in {\em Proceedings of the AAAI conference on artificial intelligence}, vol.~32, 2018.

\bibitem{zhao2023survey}
W.~X. Zhao, K.~Zhou, J.~Li, T.~Tang, X.~Wang, Y.~Hou, Y.~Min, B.~Zhang, J.~Zhang, Z.~Dong, {\em et~al.}, ``A survey of large language models,'' {\em arXiv preprint arXiv:2303.18223}, 2023.

\bibitem{wu2024beyond}
S.~Wu, X.~Tan, Z.~Wang, R.~Wang, X.~Li, and M.~Sun, ``Beyond language models: Byte models are digital world simulators,'' {\em arXiv preprint arXiv:2402.19155}, 2024.

\bibitem{wang2024mambabyte}
J.~Wang, T.~Gangavarapu, J.~N. Yan, and A.~M. Rush, ``Mambabyte: Token-free selective state space model,'' {\em arXiv preprint arXiv:2401.13660}, 2024.

\end{thebibliography}

\begin{IEEEbiography}[{\includegraphics[width=1in,height=1.25in,clip,keepaspectratio]{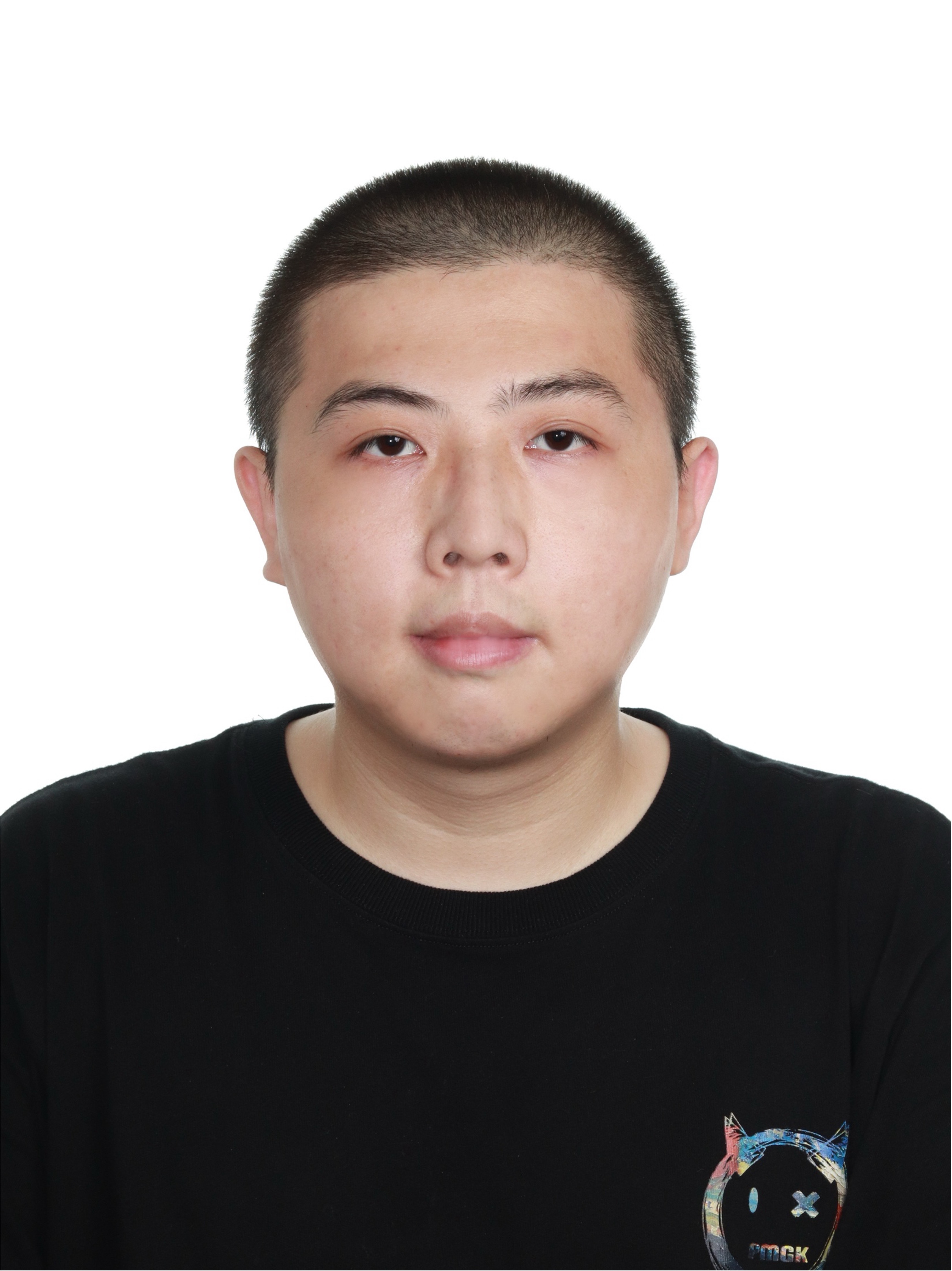}}]{Wenyang Liu} (Student Member, IEEE) received BEng and MEng degrees in the College of Computer Science from Chongqing University, Chongqing, China, in 2017 and 2020, respectively. He is currently pursuing the Ph.D. degree in the School of Electrical and Electronic Engineering from Nanyang Technological University, Singapore. His research interests include computer vision and digital forensics, with a focus on image restoration, super-resolution, and generative models.
\end{IEEEbiography}

\begin{IEEEbiography}[{\includegraphics[width=1in,height=1.25in,clip,keepaspectratio]{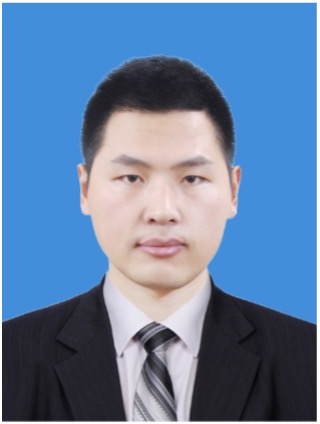}}]{Kejun Wu} (Senior Member, IEEE) received the Ph.D. degree in Information and Communication Engineering, Huazhong University of Science and Technology, Wuhan, China, in 2022. He has worked as a Research Fellow in the School of Electrical and Electronic Engineering, Nanyang Technological University, Singapore from 2022 to 2024. He is currently a Lecturer in the School of Electronic Information and Communications, Huazhong University of Science and Technology. His research interests include generative artificial intelligence, vision large language models, image restoration, and video compression, etc. He has published over 40 peer-reviewed papers including IEEE TMM, TCSVT, TOMM, OE, NeurIPS, etc. He has served as Session Chair in international conferences ICASSP 2024, IEEE ISCAS 2024, IEEE MMSP 2023, and Lead Guest Editor in JVCI.   
\end{IEEEbiography}

\begin{IEEEbiography}[{\includegraphics[width=1in,height=1.25in,clip,keepaspectratio]{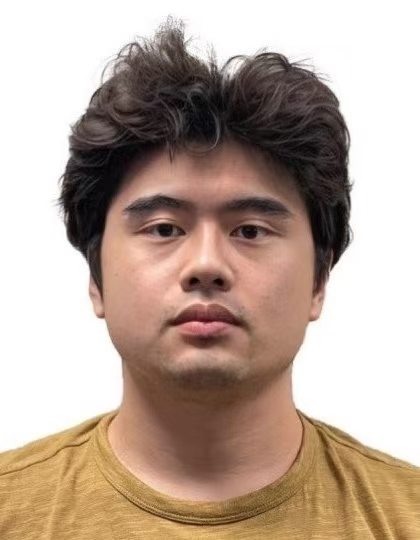}}]{Tianyi Liu} (Student Member, IEEE) received the dual BEng degree from the  University of Electronic Science and Technology of China and the University of Glasgow, in 2019, and the MSc degree from Nanyang Technological University, in 2020. He was a Graduate Research Student at The University of Tokyo from 2020 to 2022. He is currently pursuing the Ph.D.  degree in the School of Electrical and Electronic Engineering from Nanyang Technological University, Singapore. His research interests include computer vision and human-computer interaction.

\end{IEEEbiography}

\begin{IEEEbiography}[{\includegraphics[width=1in,height=1.25in,clip,keepaspectratio]{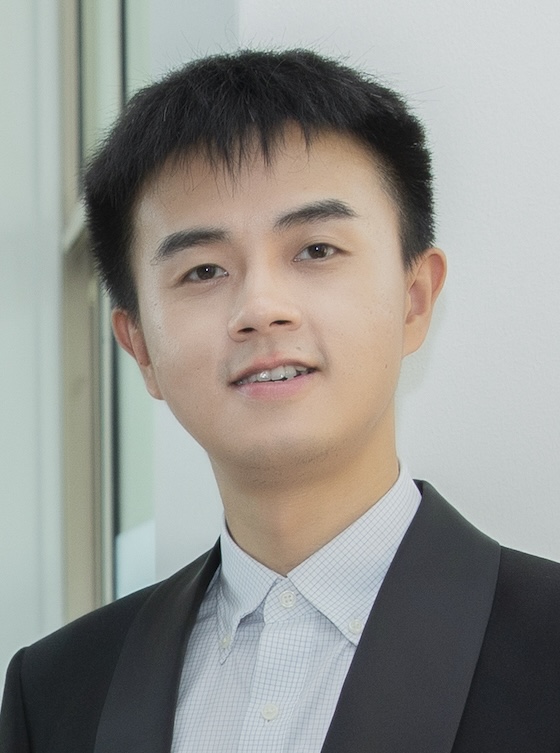}}]{Yi Wang}
(Member, IEEE) received BEng degree in electronic information engineering and MEng degree in information and signal processing from the School of Electronics and Information, Northwestern Polytechnical University, Xi’an, China, in 2013 and 2016, respectively. He earned PhD in the School of Electrical and Electronic Engineering from Nanyang Technological University, Singapore, in 2021. He is currently a Research Assistant Professor at the Department of Electrical and Electronic Engineering, The Hong Kong Polytechnic University, Hong Kong. His research interests include Image/Video Processing, Computer Vision, Intelligent Transport Systems, and Digital Forensics.
\end{IEEEbiography}

\begin{IEEEbiography}[{\includegraphics[width=1in,height=1.5in,clip,keepaspectratio]{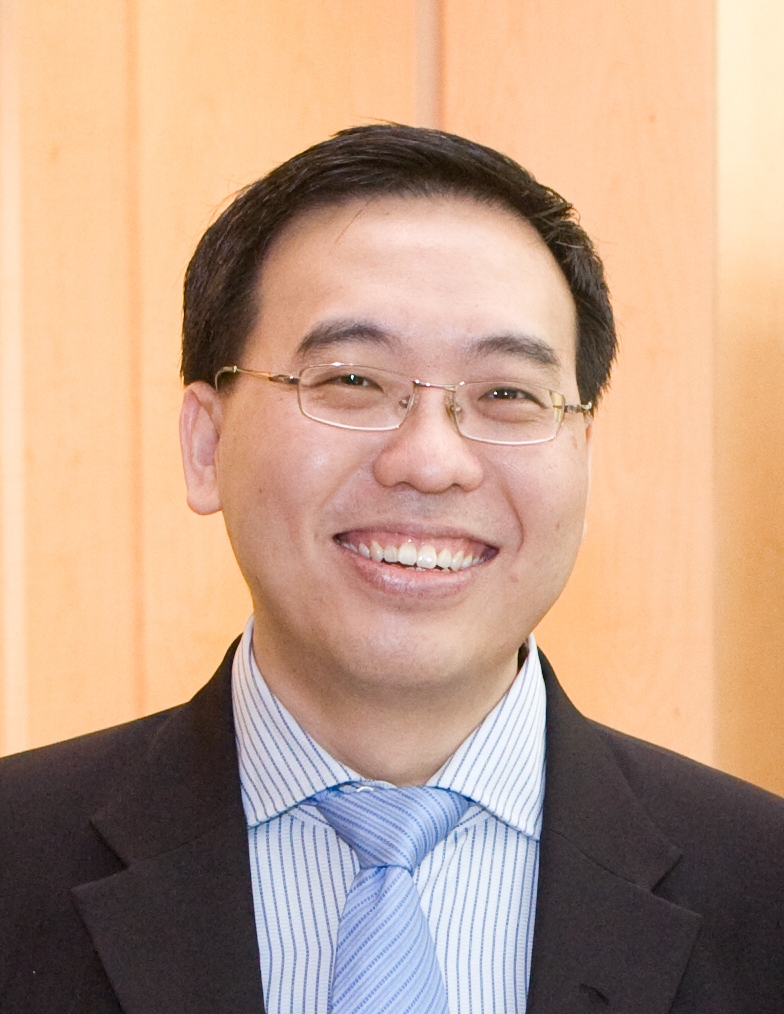}}]{Kim-Hui Yap} (Senior Member, IEEE) received the BEng and Ph.D. degrees in electrical engineering from the University of Sydney, Australia. He is currently an Associate Professor with the School of Electrical and Electronic Engineering, Nanyang Technological University, Singapore. He has authored more than 100 technical publications in various international peer-reviewed journals, conference proceedings, and book chapters. He has also authored a book titled \textit{Adaptive Image Processing: A Computational Intelligence Perspective} (Second Edition, CRCPress). His current research interests include artificial intelligence, data analytics, image/video processing, and computer vision. He has participated in the organization of various international conferences, serving in different capacities, including the technical program co-chair, the finance chair, and the publication chair in these conferences. He was an associate editor and an editorial board member of several international journals.
\end{IEEEbiography}

\begin{IEEEbiography}[{\includegraphics[width=1in,height=1.5in,clip,keepaspectratio]{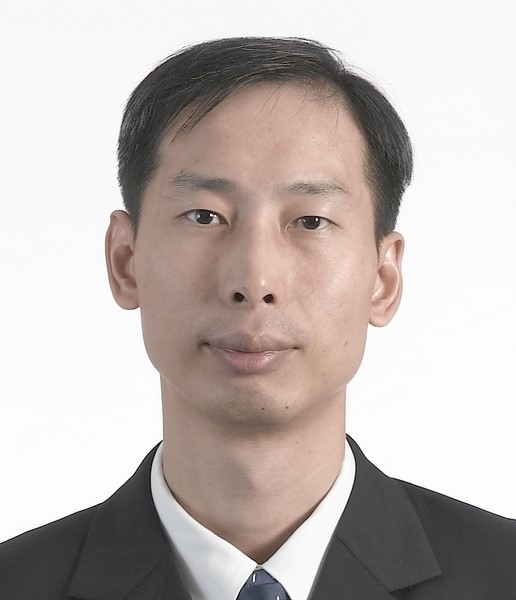}}]{Lap-Pui Chau}
(Fellow, IEEE) received a Ph.D. degree from The Hong Kong Polytechnic University in 1997. He was with the School of Electrical and Electronic Engineering, Nanyang Technological University from 1997 to 2022. He is currently a Professor in the Department of Electrical and Electronic Engineering, The Hong Kong Polytechnic University. His current research interests include image and video analytics, and autonomous driving. He is an IEEE Fellow. He was the chair of Technical Committee on Circuits Systems for Communications of IEEE Circuits and Systems Society from 2010 to 2012. He was general chairs and program chairs for some international conferences.
\end{IEEEbiography}

\end{document}